\documentclass[10pt,twocolumn,letterpaper]{article}

\usepackage{cvpr}              

\usepackage[dvipsnames]{xcolor}

\newcommand{\SynthEgo}{\emph{SynthEgo}}

\definecolor{cvprblue}{rgb}{0.21,0.49,0.74}
\usepackage{graphicx}
\usepackage{amsmath}
\usepackage{amssymb}
\usepackage{bbding}
\usepackage{color}
\usepackage{booktabs}
\usepackage{multirow}
\usepackage{pgfplots}
\usepackage{float}
\usepackage{comment}

\pgfplotsset{width=10cm,compat=1.9}
\DeclareMathOperator{\tr}{tr}
\DeclareMathOperator{\Fs}{\mathbf{F}}
\DeclareMathOperator{\Rot}{\mathbf{R}}
\DeclareMathOperator{\XX}{\mathbf{X}}

\usepackage[pagebackref,breaklinks,colorlinks,citecolor=cvprblue]{hyperref}

\def\paperID{196} 
\def\confName{3DV\xspace}
\def\confYear{2024\xspace}

\title{{SimpleEgo}: Predicting Probabilistic Body Pose from Egocentric Cameras}

\author{
Hanz Cuevas-Velasquez\thanks{Work conducted while at Microsoft Mesh Labs.}\\
Max Planck Institute for Intelligent Systems\\
Tuebingen, Germany\\
{\tt\small hanz.cuevas@tuebingen.mpg.de}
\and
Charlie Hewitt, Sadegh Aliakbarian, Tadas Baltru\v{s}aitis\\
Microsoft Mesh Labs\\
Cambridge, UK\\
{\tt\small \{chewitt,saliakbarian,tabaltru\}@microsoft.com}
}
\begin{document}
\maketitle
\begin{abstract}
    Our work addresses the problem of egocentric human pose estimation from downwards-facing cameras on head-mounted devices (HMD).
    This presents a challenging scenario, as parts of the body often fall outside of the image or are occluded.
    Previous solutions minimize this problem by using fish-eye camera lenses to capture a wider view, but these can present hardware design issues.
    They also predict 2D heat-maps per joint and lift them to 3D space to deal with self-occlusions, but this requires large network architectures which are impractical to deploy on resource-constrained HMDs. 
    We predict pose from images captured with conventional rectilinear camera lenses.
    This resolves hardware design issues, but means body parts are often out of frame.
    As such, we directly regress probabilistic joint rotations represented as matrix Fisher distributions for a parameterized body model.
    This allows us to quantify pose uncertainties and explain out-of-frame or occluded joints.
    This also removes the need to compute 2D heat-maps and allows for simplified DNN architectures which require less compute. 
    Given the lack of egocentric datasets using rectilinear camera lenses, we introduce the \SynthEgo{} dataset\footnote{Available at {\scriptsize \url{https://microsoft.github.io/SimpleEgo}}.}, a synthetic dataset with 60K stereo images containing high diversity of pose, shape, clothing and skin tone.
    Our approach achieves state-of-the-art results for this challenging configuration, reducing mean per-joint position error by 23\% overall and 58\% for the lower body.
    Our architecture also has eight times fewer parameters and runs twice as fast as the current state-of-the-art.
    Experiments show that training on our synthetic dataset leads to good generalization to real world images without fine-tuning.
\end{abstract}    
\section{Introduction}
\label{sec:intro}

Head-mounted devices (HMD) and virtual presence have become increasingly widespread in recent years and are likely to play a key role in future virtual interactions.
In these scenarios, accurate representation of the user's appearance \emph{and motion} are key to creating a sense of embodiment for the user themselves, and immersion for other users interacting with them~\cite{ferstl2021human,oh2018systematic}.

Current HMDs typically provide only very sparse input (\eg, head and hand location and orientation) and there has been significant work in producing plausible body pose from these~\cite{winkler2022questsim,aliakbarian2022flag,choutas2021learning,jiang2022avatarposer,ponton2022mmvr}.
These methods particularly struggle with generating plausible lower body poses for which almost no signal is available, leading to floating upper body avatars in existing products.
One potential solution to this problem is to use a downwards-facing camera on the HMD (\ie, an \emph{egocentric} viewpoint) to capture visual information about the entire body. 
However, even given this information-rich additional input, accurately recovering complete body pose is still a challenging problem due to partial visibility and self-occlusion.

\begin{figure}
    \centering
    \includegraphics[width=\linewidth]{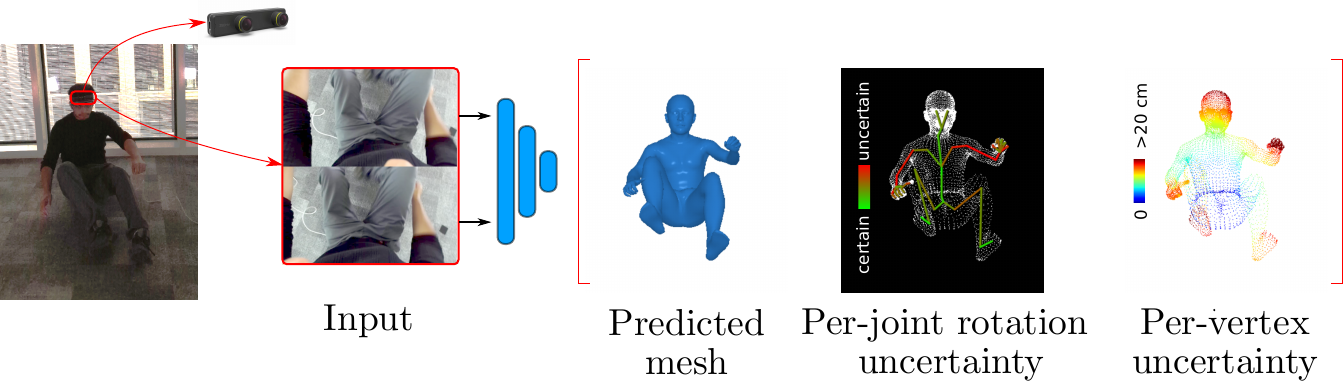}
    \caption{Our proposed method takes images from a head-mounted camera as input, and returns the parameters of a Fisher distribution for matrix rotations for each joint. From the output, we can determine joint rotations and so reconstruct the body of the user, as well as explain the pose prediction based on the uncertainties of the predicted distributions.}
    \label{fig:overview}
\end{figure}

Recently, great progress has been made in improving pose estimation in egocentric scenarios~\cite{xu2019mo, tome2019xr, akada2022unrealego}, however, there exist fundamental challenges yet to be addressed:

\textbf{(1)} 
Existing solutions for predicting full-body pose from egocentric images~\cite{xu2019mo, tome2019xr, akada2022unrealego} typically predict only 3D skeletal joint locations, which are not well-suited to directly drive an avatar. 
Although one can build on top of such approaches and fit a parametric model to predicted 3D points, such solutions are often slow, prone to error, and require a lot of manual tuning to achieve good results~\cite{Bogo:ECCV:2016,SMPL-X:2019}.
Training models to predict pose parameters rather than joint locations is also made difficult by the lack of available datasets with suitable annotations.

\textbf{(2)} 
Current methods' common practice is to use 2D heat-maps within DNN architectures to represent 2D joint locations in images~\cite{xu2019mo, tome2019xr, akada2022unrealego}.
We argue that such intermediate representations often deal poorly with joints that fall outside of the image, limiting the usability in real-world applications where robustness to such scenarios is essential.
Additionally, methods relying on 2D heat-maps have relatively deep and complex architectures, leading to larger memory consumption and slower execution, neither of which are suitable for deployment on an HMD with tight resource constraints in an immersive experience where humans are highly perceptive of latency in body motion~\cite{elbamby2018toward}.

\textbf{(3)} 
Another challenge of egocentric pose estimation is partial body visibility due to the camera placement on the HMD. 
To alleviate the effect of partial visibility on performance, existing solution typically use fish-eye lenses to capture a wider view~\cite{xu2019mo, tome2019xr, akada2022unrealego}. This results in the luxury of observing the entire body under the majority of poses, the only exception being self-occlusion.
However, fish-eye lenses are not well-suited for HMDs as they require protrusions from the form-factor of the device. 
This is not aesthetically pleasing and can cause lenses to be easily scratched or damaged.
Furthermore, images from fish-eye lenses are highly distorted, affecting the ability of deep neural networks (DNNs) to learn information about their content, including direct regression of body pose~\cite{xu2019mo}, and limiting the usage of pre-trained pose priors~\cite{zhang2021egobody}.
Lenses with a narrower field-of-view solve these issues, but also introduce their own; primarily that joints are much more likely to fall outside of the image.

To address the aforementioned challenges, we propose regression of probabilistic full-body pose parameters from egocentric images captured using a head-mounted camera with a conventional rectilinear lens. 
In particular,

\textbf{(1)} 
We propose directly predicting joint rotations for a parameterized body representation, eliminating the need for an iterative fitting process or manual tuning.

\textbf{(2)} 
We eliminate the need for heat-maps by predicting joint rotation distributions directly from input images. 
Particularly, we represent each joint rotation as a rotation matrix and estimate its probability density function as a matrix Fisher distribution~\cite{lee2018bayesian, mohlin2020probabilistic, sengupta2021hierarchical}. 
As a result, in addition to predicting the pose with joint rotations, our model is capable of predicting reliable confidence scores. 
We observe that predicting joint rotations directly allows us to use much more compact neural networks, resulting in a faster method that achieves state-of-the-art performance.

\textbf{(3)} 
Finally, 
we introduce a synthetic dataset of egocentric images using a pinhole camera model including annotations for full-body pose as well as 3D joint locations.
We are therefore able to tackle the more challenging problem of partial visibility by introducing a method that is more robust, rather than relying on easier but less plausible camera setups.
We also collect a dataset of real images to evaluate the generalization of our approach to real-world scenarios.

Our method significantly outperforms existing solutions, reducing mean per-joint position error (MPJPE) by 23\% overall, and by 58\% for the lower body. 
Further evaluations on our real test set also reveal the effectiveness of our approach in robustness to domain shift between real and synthetic data, outperforming existing baselines by a large margin in both monocular and stereo scenarios.
In addition, our monocular model runs at 100 fps on an NVIDIA 1080 GPU, over $2\times$ faster than most existing methods, while having over $8\times$ fewer parameters. 

In summary, our contributions are:
\begin{itemize}
    \item Probabilistic joint rotation prediction over the $SO(3)$ group for end-to-end egocentric pose estimation.
    \item In-depth analysis of our joint rotation uncertainty estimation; we demonstrate a strong correlation between predicted uncertainty and error measures, indicating that these predicted confidences are reliable and thus useful for various downstream tasks.
    \item State-of-the-art performance on monocular and stereo egocentric pose estimation with a comparatively simple network architecture and significantly faster execution time, providing a more realistic prospect of real-world deployment.
    \item The \SynthEgo{} dataset, a large synthetic dataset for egocentric pose estimation composed of 60K stereo pairs with body and hand pose annotations and experimental verification that our dataset can generalize well to in-the-wild images.
\end{itemize}

\section{Related work}

Human body pose prediction is a well-established field, but has typically focused on external viewpoints, with egocentric methods only appearing in recent years. 
As such, we limit our review of related work to only the most closely related external-view methods, egocentric methods, and uses of synthetic data for similar applications.

\paragraph{External view pose regression.} 
There is a large body of work tackling estimation of 3D pose from external viewpoints~\cite{chen2020monocular}. 
These can be divided into model-free and model-based methods. 
Model-free methods do not use human body models and predict 3D joint locations directly~\cite{li20143d, li2015maximum, tekin2016structured}. 
Some approaches learn to lift 2D poses and heat-maps to 3D space~\cite{martinez2017simple, li2019generating, qammaz2019mocapnet}. 
Others add constraints based on the structure of the human body to improve predictions~\cite{sun2017compositional, tekin2016structured}. 
Model-based methods output the parameters of a human model~\cite{choutas2020monocular, sengupta2021hierarchical, pavlakos2018learning, omran2018neural, varol2018bodynet, kolotouros2019learning, kolotouros2021probabilistic}. 

Following the categorization of external view approaches, current egocentric pose estimation methods~\cite{akada2022unrealego,tome2019xr,xu2019mo} would be considered model-free, with the major downside of only recovering the 3D skeleton of the user, limiting their real-world applications. 
In contrast, our approach is model-based, which allows us to recover a 3D human mesh using the predicted pose.

Due to the small number of large public datasets with 3D pose annotations, some methods rely on intermediate predictions and constraints, like predicting 2D landmarks and body segmentation~\cite{pavlakos2018learning,omran2018neural,sengupta2020synthetic} or performing edge detection~\cite{sengupta2021hierarchical}, minimizing the re-projection error of the 3D body model onto the 2D image~\cite{sengupta2021hierarchical,sengupta2020synthetic,kolotouros2019learning}, or estimating the body shape and rejection sampling poses to improve the final body mesh~\cite{sengupta2021hierarchical}. 

Intermediate predictions like 2D landmarks and segmentations allow these works to generate synthetic data easily without issues of domain gap~\cite{sengupta2020synthetic,sengupta2021hierarchical}.
However, these secondary inputs aren't useful in the egocentric scenario where the assumption that the majority of the body is inside the image frame fails~\cite{tome2019xr}.
To address this, we directly regress pose from image data without intermediate predictions, enabled by generation of highly realistic synthetic data with minimal domain gap.

Direct regression also simplifies the model architecture and reduces run-time compared to recent external-view methods, this is valuable given the performance constraints of the egocentric scenario due to limited on-board compute in HMDs.
Additionally, we inject information on the kinematic tree into our network implicitly through an extra loss term (\cref{eq:joint_regressor}) which allows for simplification of our network compared to complex recursive architectures used to achieve this explicitly~\cite{sengupta2021hierarchical}.

\paragraph{Egocentric pose estimation.} 
Current approaches for egocentric pose estimation find the 3D position of the joints by implementing a two-stage network architecture~\cite{xu2019mo,akada2022unrealego,tome2019xr}. 
The first stage predicts 2D heat-maps as a way to capture the uncertainty of each joint. 
The second stage uses the heat-maps as input, encodes them into an embedding with reduced dimension and uses a decoder to predict the 3D joint positions. 
Training these types of networks is expensive as they require $N \times M$ heat-maps per joint as well as large encoder-decoder models. 
These architectures also introduce extra hyper-parameters that must be tuned, like the kernel size of the heat-maps and the dimensionality of the embedding.
Finally, although the 2D heat-maps can capture the uncertainty of self-occluded joints~\cite{tome2019xr}, they lose the information for out-of-frame joints.

Unlike previous egocentric methods, we predict joint rotations for a human parametric model distributed according to a matrix Fisher distribution, following the external-view approach of \citet{sengupta2021hierarchical}. 
The rotation distribution allows our network to model the uncertainty of the joints, even when they are outside the frame, as well as reconstruct the body of the user and not only the skeleton. 
As our approach does not need a 2D-to-3D lifting step, it can easily be trained end-to-end and enables a simpler and faster model.

\paragraph{Synthetic egocentric data.} 
Collecting a real dataset for egocentric pose estimation while capturing diversity in pose, shape, and environment is difficult and expensive. 
Consequently, current methods render synthetic datasets to train their networks~\cite{xu2019mo, tome2019xr, akada2022unrealego}. 

Mo$^2$Cap$^2$~\cite{xu2019mo} is the first egocentric dataset to be released. 
The authors use a single fish-eye lens camera attached to the head of the user, and generate synthetic humans by randomly posing SMPL~\cite{loper2015smpl} models on real backgrounds. 
The dataset includes 15 landmarks per image.

xR-EgoPose~\cite{tome2019xr} provides a more visually realistic dataset where the lighting of the characters matches the background. 
This dataset includes 25 body and 40 hand landmarks and heat-maps, but only 16 body landmarks are used during training and inference. 
The dataset includes simple every-day human motions where most of the motions start from rest pose and again uses a fish-eye lens. 

UnrealEgo~\cite{akada2022unrealego} captures a higher variety of poses from a stereo fish-eye camera pair. 
Similar to the previous method, the dataset includes hand and body joint positions but only uses 16 during training and inference. 

In contrast, the \SynthEgo{} dataset includes full-body pose parameters including hands, giving a total of 54 joints, all of which are used during training and inference.
We render stereo pairs using a pinhole camera model motivated by physical hardware design constraints, this provides a more challenging and realistic scenario where parts of the body can be out of frame.
Our dataset also incorporates a huge variety of body shapes and poses, clothing and environments with a high level of visual realism.


\section{Method}

In this section we present our method.
We describe how we use the matrix Fisher distribution~\cite{lee2018bayesian, mohlin2020probabilistic} to represent rotations and give details of the loss functions we use for training.
Finally, we provide a summary of the human parametric model and synthetic data generation pipeline~\cite{tech_report}.

\subsection{Probabilistic joint rotation}

The goal of our method is to estimate the probability distribution over joint rotations $\Rot = \{\mathbf{R}_i\}^{N}_{i=1}$ conditioned on input image data $\XX$, $p(\Rot|\XX)$. 
We represent the rotation matrix of each joint as a random variable, $\Rot_i \in SO(3)$, distributed according to a matrix Fisher distribution parameterized by $\Fs_i \in \mathbb{R}^{3\times 3}$~\cite{lee2018bayesian,mohlin2020probabilistic,sengupta2021hierarchical}.

We train a neural network to regress Fisher parameters $\Fs = \{\mathbf{F}_i\}^{N}_{i=1}$ given input image data $\XX$.
From these predicted parameters we can calculate the expected rotation, $\mathbf{\hat{R}}_i$, and proper singular values, $\mathbf{S}_i$, for each joint~\cite{lee2018bayesian}.
The concentration for joint $i$, $\kappa_{i,j}$, around principal axis $j$ can be obtained by $\kappa_{i,j}=s_{i,k}+s_{i,l}$ for $(j,k,l) \in \{(1,2,3),(2,1,3),(3,1,2)\}$. 
The concentration parameters can be different about each principal axis, allowing us to model rotational uncertainty per-axis.
\cref{sec:uncertanty_eval} discusses how the concentration parameters can capture the degrees of freedom of each joint.

\subsection{Loss functions}

We train the neural network by minimizing loss $\mathcal{L} = \mathcal{L}_{FNLL} + \mathcal{L}_J$. 
$\mathcal{L}_{FNLL}$ is the matrix Fisher negative log-likelihood (\cref{eq:fisher_log_likelihood}), promoting accurate local joint rotations.
$\mathcal{L}_J$ supervises the 3D joint positions regressed from the SMPL-H* model (\cref{eq:joint_regressor}).
This causes the network to consider the effect of the predicted rotations on the final pose, as the positions of child joints are influenced by the rotation of their parents in the kinematic tree of our body model. 

\paragraph{Matrix Fisher negative log-likelihood.} 
We minimize the negative log-likelihood (NLL) of the ground-truth rotation matrices, $\Rot$, given the estimated Fisher distributions, $\Fs$, for all joints.
The objective function is defined as by \citet{mohlin2020probabilistic} for $N$ joints:
\begin{equation} \label{eq:fisher_log_likelihood}
 \begin{aligned}
 \mathcal{L}_{FNLL}&=\sum_{i=1}^{N}log(c(\Fs_i))-\tr(\Fs_i^\top \Rot_i)  
 \end{aligned}
\end{equation}

\paragraph{SMPL-H* joint loss.} 
The differentiable $\textit{SMPL-H*}$ model, and joint regressor, $\mathcal{J}$, described in \cref{sec:body_model}, are used to regress the 3D joint locations from the means of the predicted matrix Fisher distributions, $\hat{\Rot}$, and ground-truth joint rotation matrices, $\Rot$, using the ground-truth body shape, $\boldsymbol\beta$.
An L2 loss is then used to minimize the distance between the predicted and ground-truth joint locations:
\begin{equation} \label{eq:joint_regressor}
 \begin{aligned}
J_{3D}(\Rot,\boldsymbol\beta)=&\mathcal{J}(\textit{SMPL-H*}(\Rot, \boldsymbol\beta))\\
\mathcal{L}_{J}=&\left \| J_{3D}(\hat{\Rot},\boldsymbol\beta)- J_{3D}(\Rot,\boldsymbol\beta) \right \|^2_2
\end{aligned}
\end{equation}

For exploration of additional loss functions see the supplementary material.

\subsection{Parametric human model}
\label{sec:body_model}

To generate the images and ground-truth data in our dataset we use the parametric human model described in~\cite{tech_report}.
The body shape is defined by parameters $\boldsymbol\beta\in \mathbb{R}^{10}$, identical to SMPL-H~\cite{romero2022embodied}.
The body pose is defined by the relative 3D rotation of the bones formed by $N = 54$ joints, including the root joint.
The first 52 components are identical to SMPL-H, with the final two components representing the eyes.
In SMPL-H these joint rotations are represented as axis-angles, but here we choose instead to represent them as matrices. 
For simplicity we refer to this model as SMPL-H*, indicating SMPL-H with the addition of two eye joints.

SMPL-H* can be interpreted as a differentiable function that maps the input pose and shape parameters to an output mesh $\mathbf{V} \in \mathbb{R}^{M\times 3}$ with $M$ vertices.
The 3D joint locations, $J^{3D}$, can be obtained using a linear regressor, $J^{3D} = \mathcal{J}(\mathbf{V})$ where $\mathcal{J} \in \mathbb{R}^{N \times M}$.

\subsection{Synthetic dataset}

We use the SMPL-H* parametric body model to create a synthetic dataset with ground-truth pose and shape parameters, as well as joint locations.
This allows us to compare our method with recent work which only provides 3D joint \emph{position} annotations~\cite{tome2019xr,akada2022unrealego}.
The \SynthEgo{} dataset is available to download from the project website: {\small \url{https://microsoft.github.io/SimpleEgo}}.

To construct the \SynthEgo{} dataset we render 60K stereo pairs at $1280 \times 720$ pixel resolution, building on the pipeline of \citet{tech_report}.
This dataset is comprised of 6000 unique identities, each performing 10 different poses in 10 different lighting environments.
Each identity is made up of a randomly sampled body shape, skin textures sampled from a library of 25 and randomly recolored, and clothing assets sampled from a library of 202.
Lighting environments are sampled from a library of 489 HDRIs, to ensure correct disparity of the environment between the stereo pair, we project the HDRI background onto the ground plane.
Poses are sampled from a library of over 2 million unique poses and randomly mirrored; sampling is weighted by the mean absolute joint angle and common poses like T-pose are significantly down-weighted to increase diversity.
A comparison to other datasets can be seen in \cref{tab:dataset_comp}.

\begin{table}
\centering
\resizebox{\linewidth}{!}{%
\begin{tabular}{lcccc} 
\toprule
& Mo$^2$Cap$^2$~\cite{xu2019mo} & xR-EgoPose~\cite{tome2019xr} & UnrealEgo~\cite{akada2022unrealego} & \SynthEgo{} (ours)  \\ 
\midrule
Unique Identities  & 700          & 46            & 17            &  6000        \\
Environments       & Unspecified  & Unspecified   & 14            &  489         \\
Body Model         & SMPL         & Unspecified   & UnrealEngine  &  SMPL-H*     \\
Lens Type          & Fisheye      & Fisheye       & Fisheye       &  Rectilinear \\
Mono/Stereo        & Mono         & Mono          & Stereo        &  Stereo      \\
Body Shape GT      &              &               &               &  \checkmark  \\
Joint Location GT  & \checkmark   & \checkmark    & \checkmark    &  \checkmark  \\
Joint Rotation GT  &              &               &               &  \checkmark  \\
Realism            & Low          & Medium        & High          &  High        \\
\bottomrule
\end{tabular}%
}
\caption{Comparison of synthetic egocentric pose datasets.}
\label{tab:dataset_comp}
\end{table}

\begin{figure}
    \centering
     \includegraphics[width=0.32\linewidth]{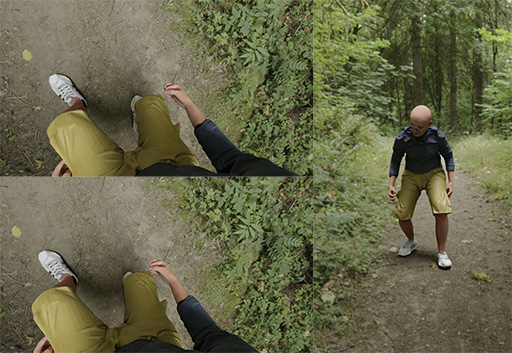}
     \includegraphics[width=0.32\linewidth]{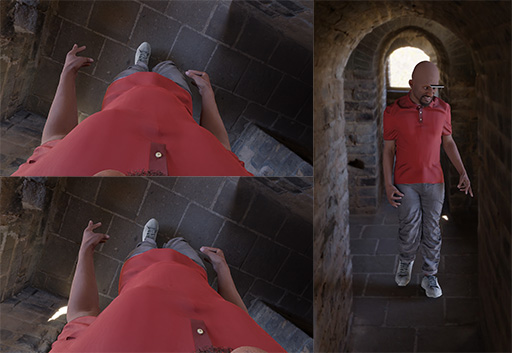}
     \includegraphics[width=0.32\linewidth]{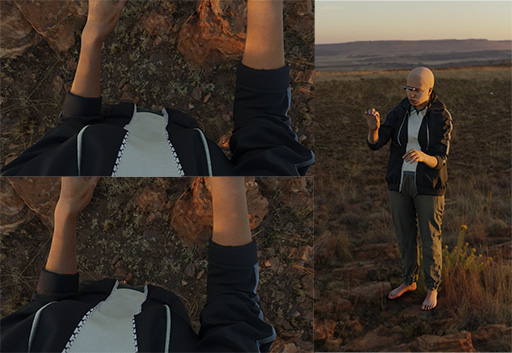}
    \caption{Example scenes from the \emph{SynthEgo} dataset showing the left and right egocentric views used for training and an external viewpoint used for visualization only.}
    \label{fig:synthetic_data}
\end{figure}

We position the camera on the front of the forehead looking down at the body. 
The camera uses a pinhole model approximating the ZED mini stereo\footnote{\url{https://www.stereolabs.com/}}.
We add uniform noise within $\pm1$ cm to the location and $\pm10^\circ$ around all axes of rotation of the camera to simulate misplacement and movement of the HMD on the head. 
The resulting images are typically quite challenging for pose estimation, as many parts of the body are often not seen by the camera as seen in example images from the dataset in \cref{fig:synthetic_data}.

\section{Experiments}
This section describes the training process and compares our probabilistic method with the state-of-the-art on our synthetic and real datasets. 
We also provide an interpretation of the predicted rotation uncertainties and a brief summary of our method's performance.

\subsection{Training details}
\label{sec:train_details}
For all experiments, we split the \SynthEgo{} dataset into 45K frames for training, 5K frames for validation, and 10K frames for testing. 
The left and right images are resized to $480 \times 270$ pixels. 
Following \citet{akada2022unrealego}, we use the left image to train the monocular network and both to train the stereo. 
For data augmentation we include blurs, modulations to brightness and contrast, addition of noise, and conversion to grayscale to bridge the domain gap between synthetic and real data~\cite{wood2021fake}. 

We use a ResNet-34~\cite{he2016deep} backbone, the resulting feature vector is passed to a linear layer before predicting the Fisher parameters for each joint.
In the stereo case we use a Siamese network~\cite{koch2015siamese} with shared weights. 
To combine the features of the two streams we utilize late fusion by simply concatenating the features output by each branch.
We train for 200 epochs using the Adam optimizer~\cite{kingma2014adam} with a learning rate of $5e-4$ and batch size of 128.

We retrain xR-EgoPose~\cite{tome2019xr} and UnrealEgo~\cite{akada2022unrealego} on the \SynthEgo{} dataset using the same image size and data augmentation pipeline as described above. 
Since the source code of xR-EgoPose is not available, we re-implement it based on the paper description~\cite{tome2019xr}. 
Following \citet{tome2019xr}, we create heat-maps with a kernel size of $\sigma=10$ and down-sample the heat-maps four times the input size to $120\times72$ pixels. 
Similar to our monocular and stereo networks, xR-EgoPose is trained using the left image and UnrealEgo using the stereo pair. 
The heat-map networks are trained for 200 epochs, and the 3D landmark network and heat-map reconstruction are trained for another 200 epochs. 

\subsection{Pose Prediction}

To the best of our knowledge there is no work that estimates joint \emph{rotations} directly using images from head-mounted cameras, existing works instead predict 3D joint \emph{locations}~\cite{tome2019xr,akada2022unrealego}.
To compare our approach with these works, we use the SMPL-H* model to regress 3D joint locations from predicted rotations (\cref{eq:joint_regressor}), considering three cases for obtaining body shape, $\beta$. 
(1) The body shape is obtained from an enrollment step, which can give us a value equivalent to the ground truth, $\beta_{gt}$. 
(2) The network is modified to predict the body shape, $\beta_{pred}$, see supp. materials for details. 
(3) A default body shape, $\beta_{def}$, is used.

Mean per joint position error (MPJPE), and Procrustes analysis mean per joint position error (PA-MPJPE) are used for evaluation.
Following previous approaches, Procrustes analysis is conducted excluding the hand joints~\cite{xu2019mo,akada2022unrealego}. 
Both metrics are used to evaluate the error on the upper body, lower body, and hand joints separately.
XR-Egopose~\cite{tome2019xr} uses monocular input and UnrealEgo~\cite{akada2022unrealego} uses a stereo-pair.
For a fair comparison, we train a monocular and a stereo version of our approach and compare each with their respective counterpart. 
We don't compare against Egoglass~\cite{zhao2021egoglass} since UnrealEgo already outperforms it, or SelfPose~\cite{tome2020selfpose} as the model is not publicly available.

We evaluate primarily using the \SynthEgo{} dataset as it is the only existing large-scale dataset with ground-truth pose annotations to train our method, as well as providing a more challenging benchmark.
To evaluate the generalization of our approach to in-the-wild scenarios, we collect a dataset of $8$K real images and also evaluate on this.

Finally, we assess the benefit of probabilistic regression by comparing matrix Fisher NLL loss with standard L2 loss for rotation matrices.
We also compare modifications of our method to regress deterministic and probabilistic 3D joint positions on the UnrealEgo~\cite{akada2022unrealego} dataset against previous methods.

\begin{table*}
\centering
\footnotesize
\resizebox{\linewidth}{!}{%
\begin{tabular}{ll@{\hskip 0.5in}cccc@{\hskip 0.5in}cccc} 
\toprule
\multicolumn{1}{c}{\multirow{2}{*}{Input}} & \multicolumn{1}{c}{\multirow{2}{*}{Method}} & \multicolumn{4}{c}{MPJPE (mm)}                                     & \multicolumn{4}{c}{PA-MPJPE (mm)}                                   \\ 
                        & & Upper body     & Lower body     & Hands           & All             & Upper body     & Lower body     & Hands           & All             \\ 
\midrule
\multirow{5}{*}{Monocular} & xR-EgoPose~\cite{tome2019xr}              & 69.36          & 172.80         & 195.23          & 156.53          & 63.82          & 87.68          & 185.87          & 135.60          \\

& Ours (pose+$\beta_{gt}$)             & 71.28 & 80.10 & 204.09 & 146.53 & 52.38 & 67.33 & 172.96  & 121.86 \\

& Ours (pose+$\mathcal{L}_j$+$\beta_{def}$)             & 65.33 & 89.31 & 165.74 & 125.11 & 44.69 & 61.02 & 137.16 & 98.78  \\

& Ours (pose+$\mathcal{L}_j$+$\beta_{pred}$)  & 64.93 & 87.52 & 163.69 & 123.56 & 44.84 & 61.17 & 137.30 & 98.93  \\

& Ours (pose+$\mathcal{L}_j$+$\beta_{gt}$)             & \textbf{61.36} & \textbf{71.90} & \textbf{160.86} & \textbf{118.39} & \textbf{43.58} & \textbf{59.82} & \textbf{136.66} & \textbf{97.99}  \\
\midrule
\multirow{5}{*}{Stereo} & UnrealEgo~\cite{akada2022unrealego}               & 65.10          & 165.32         & 174.81          & 142.76          & 60.58          & 86.25          & 172.04          & 126.78          \\

& Ours (pose+$\beta_{gt}$) & 68.54 & 73.99 & 194.41 & 139.37 & 49.56 & 64.26 & 164.15 & 115.67  \\

& Ours (pose+$\mathcal{L}_j$+$\beta_{def}$)             & 62.08 & 86.51 & 153.80 & 117.11 & 41.74 & 58.97 & 126.49 & 91.69  \\
& Ours (pose+$\mathcal{L}_j$+$\beta_{pred}$)       & 61.35 & 83.91 & 151.33 & 115.10 & 41.95 & 58.07 & 126.82 & 91.79  \\
& Ours (pose+$\mathcal{L}_j$+$\beta_{gt}$)             & \textbf{57.95} & \textbf{68.85} & \textbf{148.38} & \textbf{110.01} & \textbf{40.50} & \textbf{57.61} & \textbf{125.70} & \textbf{91.68}  \\
\bottomrule
\end{tabular}%
}
\caption{Evaluation on the \SynthEgo{} dataset with monocular and stereo inputs. $\beta_{gt}$, $\beta_{pred}$, and $\beta_{def}$ indicate that the joint locations were regressed using the ground truth, predicted, and default body identity, respectively. $\mathcal{L}_j$ specifies that the network was trained using our joint loss. We observe that even when the ground truth body shape is not given, our model still performs better than recent approaches. We also see that the extra information the right image provides helps the model to predict more accurate joint locations.}
\label{tab:sx_dataset_eval}
\end{table*}

\paragraph{Evaluation on synthetic data.}
\cref{tab:sx_dataset_eval} shows that our method outperforms the state-of-the-art by a large margin, even when body shape is not known or predicted
We observe that incorporating our 3D joint loss, $\mathcal{L}_{3D}$, leads to an improvement in the final pose compared to a model that only optimizes local rotations. 

Overall, our stereo network has the best performance. 
From \cref{fig:synthetic_data} we observe that the extra information provided by the right image helps the network to better predict extremities. 
We also note that UnrealEgo and xR-EgoPose perform particularity poorly for lower body joints. 
This may be caused by the fact that the legs are not always visible, and that 2D heat-maps cannot provide uncertainties for joints outside of the image frame. 

\paragraph{Evaluation on real-world data.}
To evaluate the performance on real-world data, we recorded a dataset of 8378 stereo pair images from 11 different subjects performing actions like squatting, sitting, stretching, crossing arms, and interacting with small objects. 
The dataset was captured using a camera rig with three synchronized Azure Kinects (AK) arranged in a semi-circle around the subject. 
During the recording, the subject also wore a ZED mini camera mounted on an adjustable strap on their head. 
We predict 2D dense body landmarks from the RGB outputs of the AKs, and obtain the ground-truth pose and shape by fitting the SMPL-H* model to the 2D landmark observations, see \citet{tech_report} for details. 
We initialize the optimization using the method of \citet{choutas2020monocular}. 
We manually synchronize the images from the head mounted camera with the ground-truth poses and remove any frames with poor fitting results\footnote{Further information on the capture rig and examples of the data are given in the supplementary material.}.

\begin{figure}
\centering
   \includegraphics[width=\linewidth]{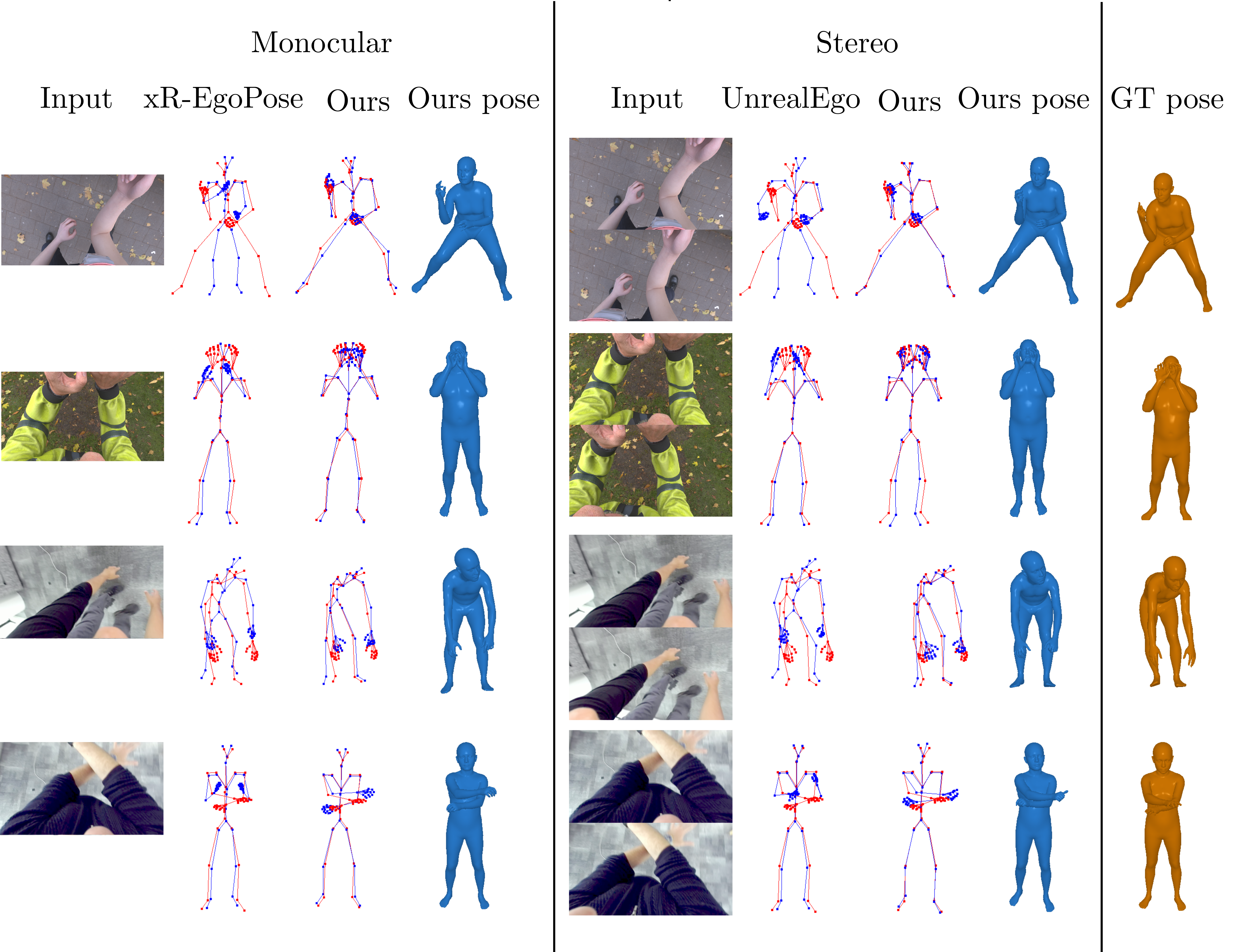}
   \vspace{-2em}
   \caption{Comparison of 3D joint location results (blue) overlaid on GT (red) for two synthetic and two real images. We also show predicted (blue) and GT (orange) body meshes. Our method accurately recovers joint locations \emph{and} rotations.}
\label{fig:output_preds}
\end{figure}

\begin{table}
\centering
\resizebox{\linewidth}{!}{%
\begin{tabular}{llcccc} 
\toprule
\multicolumn{1}{c}{\multirow{2}{*}{Input}} & \multicolumn{1}{c}{\multirow{2}{*}{Method}} & \multicolumn{4}{c}{PA-MPJPE (mm)}      \\ 
            &            & Upper body & Lower body & Hands & All  \\ 
\midrule
\multirow{2}{*}{Monocular} & xR-EgoPose~\cite{tome2019xr}   &      50.18               &    76.76             &   127.34            &   97.48       \\
                           & Ours  &      \textbf{38.48}      &    \textbf{62.35}    &    \textbf{98.94}   & \textbf{76.05} \\
\midrule
\multirow{2}{*}{Stereo} & UnrealEgo~\cite{akada2022unrealego}    &      48.06          &     77.06           &    117.85           &   91.67   \\
                        & Ours  &    \textbf{34.00}   & \textbf{54.59}      &   \textbf{87.78}    &   \textbf{67.31}   \\
\bottomrule 
\end{tabular}%
}
\caption{Comparison with previous methods for monocular and stereo evaluations on our real dataset. We significantly outperform previous methods.}
\label{tab:real_dataset_eval}
\end{table}

Since no methods estimate global translation and rotation, we report only the PA-MPJPE.
In practice, global translation and rotation would be provided by device-specific head tracking implementations.
The real dataset was only used for testing and no fine-tuning was done. 
Similar to the synthetic case, \cref{tab:real_dataset_eval} shows that our method obtains lower error than the state-of-the-art. 
Qualitative results are shown in \cref{fig:output_preds}.

\paragraph{Deterministic vs probabilistic approach.}
To assess the importance of our probabilistic formulation, we retrained our model to predict rotation matrices with matrix L2 loss~\cite{zhou2019continuity,larochelle2007distance}. 
When using L2 loss we obtain a MPJPE of 161mm and 148mm for the monocular and stereo cases.
Our probabilistic approach obtains a lower error of 147mm and 139mm, respectively. 
This consistent difference shows that training with a probabilistic rotation distribution loss provides more accurate results.

\begin{table}
\centering
\footnotesize
\resizebox{\linewidth}{!}{%
\begin{tabular}{llcc} 
\toprule
\multicolumn{1}{c}{Input} & \multicolumn{1}{c}{Method}              & MPJPE (mm) & PA-MPJPE (mm)  \\ 
\midrule
Monocular & xR-EgoPose~\cite{tome2019xr}               & 112.86             &       88.71 \\ 
Stereo & UnrealEgo~\cite{akada2022unrealego}                &  \textbf{79.06}     &          \textbf{64.65} \\

Stereo & Ours (det. 3D joints) &    118.10   &      101.93     \\
Stereo &  Ours (prob. 3D joints) &    80.00   &      67.60     \\
\bottomrule
\end{tabular}%
}
\caption{Probabilistic (prob.) vs deterministic (det.) 3D joint position estimation on the UnrealEgo dataset. Note the performance of our probabilistic approach despite having a simpler model.}
\label{tab:comp_stereo_unreal_dataset}
\end{table}

To compare against other methods directly, we modify our network to instead predict parameters $(\mu_x, \mu_y, \mu_z, \sigma)$ of a circular 3D Gaussian and train it using Gaussian negative log likelihood (GNLL) loss~\cite{kendall2017uncertainties, wood2022dense}.
We compare the models using the UnrealEgo dataset~\cite{akada2022unrealego} as it provides a large diversity of poses and people.
\cref{tab:comp_stereo_unreal_dataset} shows that probabilistic 3D joint position regression can obtain a similar performance with a significantly simpler network architecture. 

We hypothesize that learning a distribution gives more freedom to the network to discount challenging samples 
and focus on making precise predictions for visible ones, similar to L2 loss versus GNLL loss used by \citet{wood2022dense}.

\subsection{Uncertainty Estimation and Explainability}
\label{sec:uncertanty_eval}
In this section, we further evaluate the quality of the estimated uncertainties. 
Particularly, we demonstrate that not only do such uncertainty estimates capture extra information and priors about body pose, we  empirically show that the estimated uncertainties are reliable. 
While the former allows us to better explain the prediction of the model, the latter is of significant importance when it comes to deployment of our method in downstream tasks such as avatar animation, where uncertainty estimates can be used as a measure of reliance of the predicted poses.

\paragraph{Per-joint and full-body uncertainty estimation.} 
The concentration parameters predicted by our network can be used to estimate the confidence of predicted joint rotations. 
This is a key element of our model, as it lets us interpret and explain the output.

\begin{figure}
    \centering
    \includegraphics[width=\linewidth]{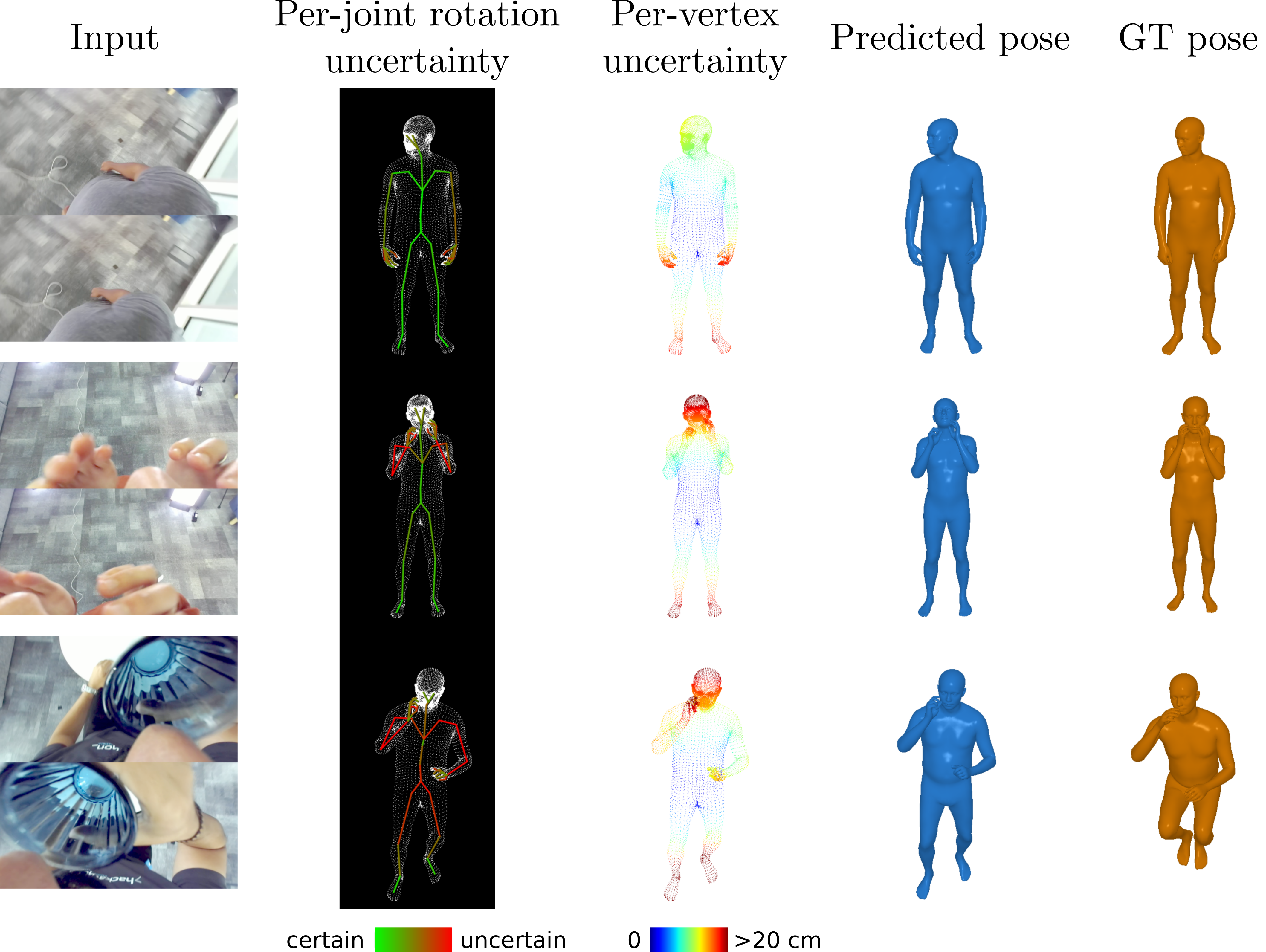}
    \caption{Uncertainty outputs of our method on our real dataset. The second column displays the uncertainty of each joint obtained by summing the concentration parameters along the three axes of each joint. Joints with high uncertainty are typically not visible in the input image. The third column shows the per-vertex uncertainty obtained by sampling the joint rotation distributions.
    }
\label{fig:uncertainty_pred}
\end{figure}

We use concentration parameters, $\kappa_{i,}$, from the predicted Fisher matrix distribution to find how certain the network is about predicted joint rotation $\Rot_i$.
To achieve this, we sum the concentration parameter of each axis, $\mathcal{K}_i=\sum^3_{j=1} \kappa_{i,j}$.
Higher values of $\mathcal{K}_i$ represent higher confidence in the predicted joint rotation, $\Rot_i$.
The second column of \cref{fig:uncertainty_pred} shows that the network learns to have low confidence for joints that are occluded or not visible in the image. 

While the predicted uncertainty estimates are local to each joint, this can be extended to the entire body by propagating the joint uncertainties through the kinematic tree of our body model. 
This is important as it allows us to evaluate the uncertainty of the entire pose, as the position of a joint is affected by everything above it in the kinematic tree. 
For example, we can generate per-vertex uncertainty which reflects the estimated variance of each body part in Euclidean space~\cite{sengupta2021hierarchical}, as shown in the third column of \cref{fig:uncertainty_pred}.

\paragraph{Body pose priors.}
Empirically, we observe that our model implicitly learned interesting priors on body pose. 
Specifically, through analysis of the uncertainty estimated for different joints in the body, we observe that our model tends to estimate lower uncertainties for  axes that a joint cannot physically rotate around.
The model has therefore learned the degrees of freedom of each joint, without being explicitly encouraged to do so.

\begin{figure}
    \centering
    \includegraphics[width=0.26\linewidth]{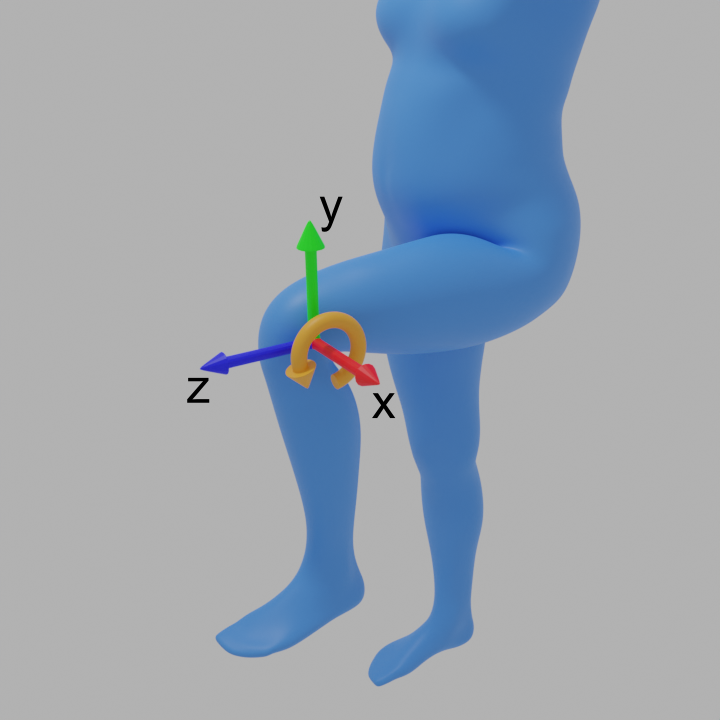}  \includegraphics[width=0.26\linewidth]{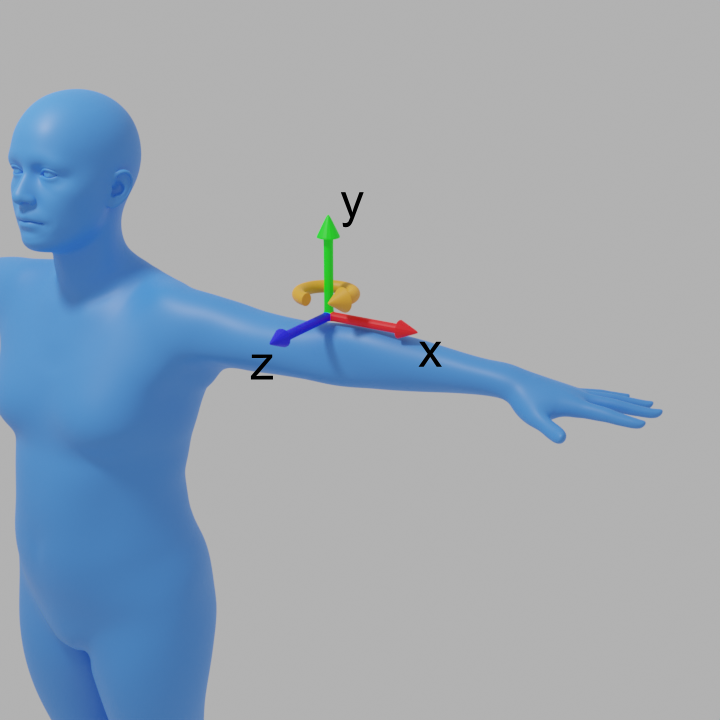}  \includegraphics[width=0.26\linewidth]{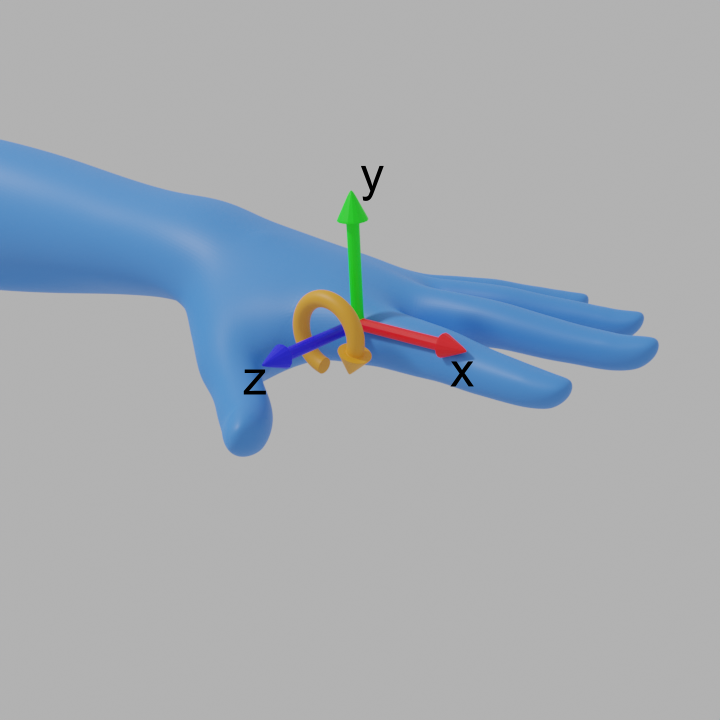} \\ \vspace{0.5em}
    \resizebox{0.8\linewidth}{!}{
        \pgfplotsset{width=0.8\textwidth, height=0.35\textwidth, compat=1.3, 
        grid style={dashed,gray}}  
        
        \begin{tikzpicture}[trim axis left, trim axis right]
        \begin{axis}[
            ybar,
            enlargelimits=0.1,
            legend style={
                at={(0.5, -0.2)},
                anchor=north,
                /tikz/every even column/.append style={column sep=0.5cm},
                font=\Large
            },
            legend columns=-1,
            ylabel={Concentration},
            y label style={at={(axis description cs:-0.02,.5)}},
            symbolic x coords={L Knee,R Knee,L Elbow,R Elbow,L Index1,R Index1},
            ymajorticks=false,
            xtick=data,
            grid=both,
            label style={font=\Large},
            tick label style={font=\Large},
        ]
        \definecolor{red_bar}{RGB}{236,153,153}
        \definecolor{green_bar}{RGB}{102,204,102}
        \definecolor{blue_bar}{RGB}{153,204,255}
        
        \addplot[black,fill = red_bar] coordinates {(L Knee,10.89452) (R Knee,10.444172) (L Elbow,11.538357) (R Elbow,11.506859) (L Index1,16.398218) (R Index1,16.296326) };
        \addplot[black,fill = green_bar] coordinates {(L Knee,15.043518) (R Knee,15.832653) (L Elbow,10.126762) (R Elbow,9.728942) (L Index1,14.566329) (R Index1,13.940974) };
        \addplot[black,fill = blue_bar] coordinates {(L Knee,17.904772) (R Knee,17.68317) (L Elbow,13.1781845) (R Elbow,12.794781) (L Index1,7.075334) (R Index1,7.129943) };
        \legend{\textit{x-axis}, \textit{y-axis}, \textit{z-axis}}
        \end{axis}
        \end{tikzpicture}
    }   
    \caption{Concentration parameter and joint degree-of-freedom relationships. Averaged concentration parameters from our real dataset show higher concentration values are correlated with axes with lower freedom.}
    \label{fig:dof_concentration}
\end{figure}
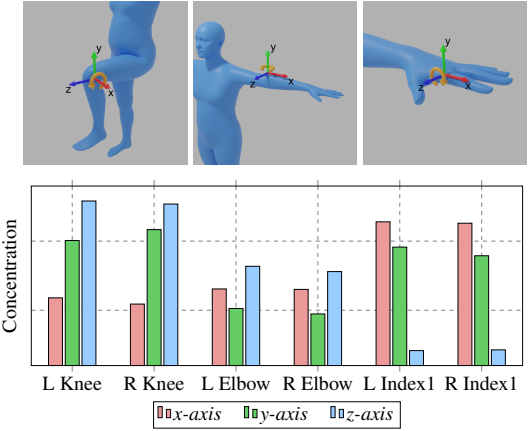

\begin{figure*}
    \centering
    \small
    \begin{tabular}{c @{} c @{} c}
        \includegraphics[trim={50px 15px 50px 50px}, clip, width=0.32\textwidth]{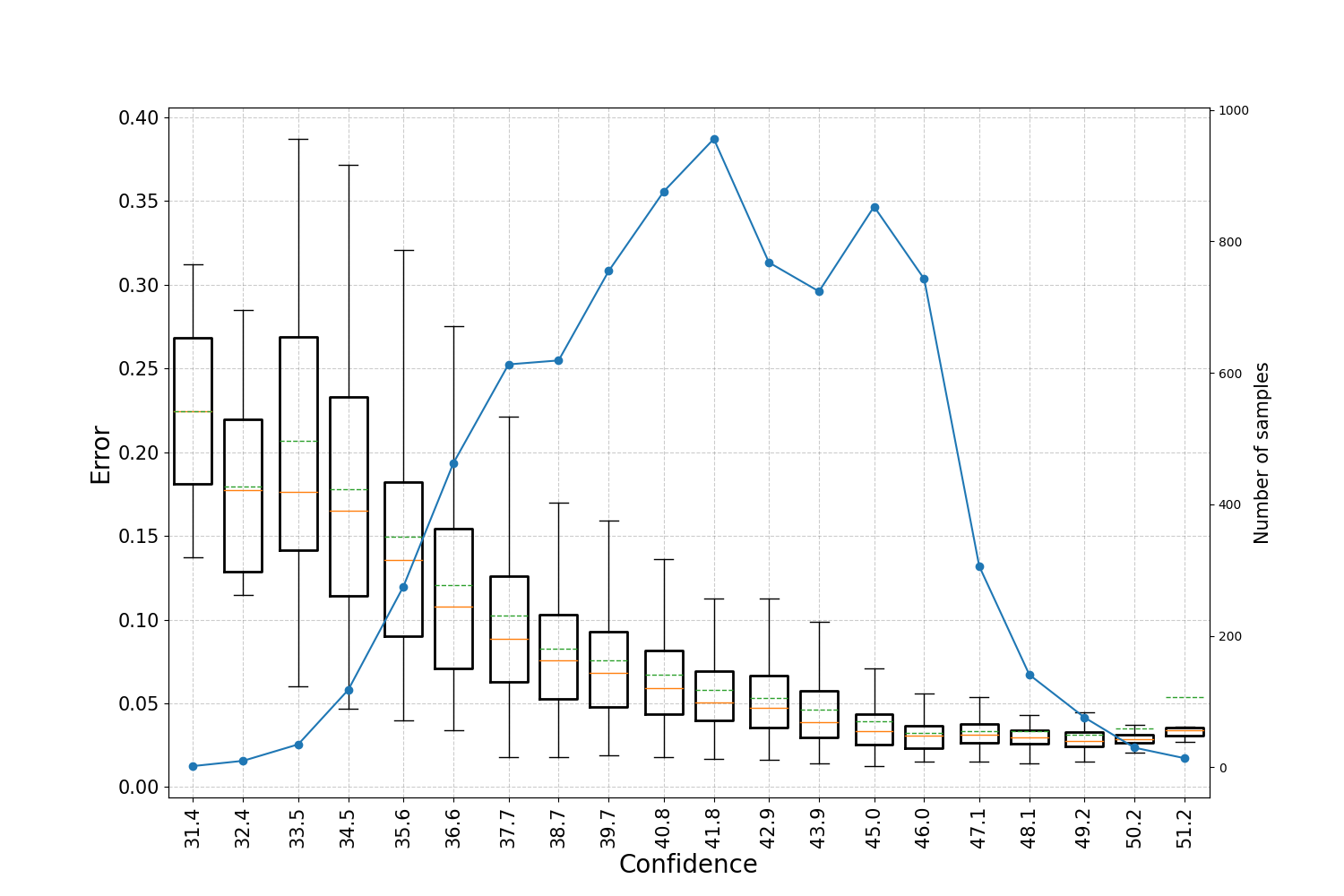} & 
        \includegraphics[trim={50px 15px 50px 50px}, clip, width=0.32\textwidth]{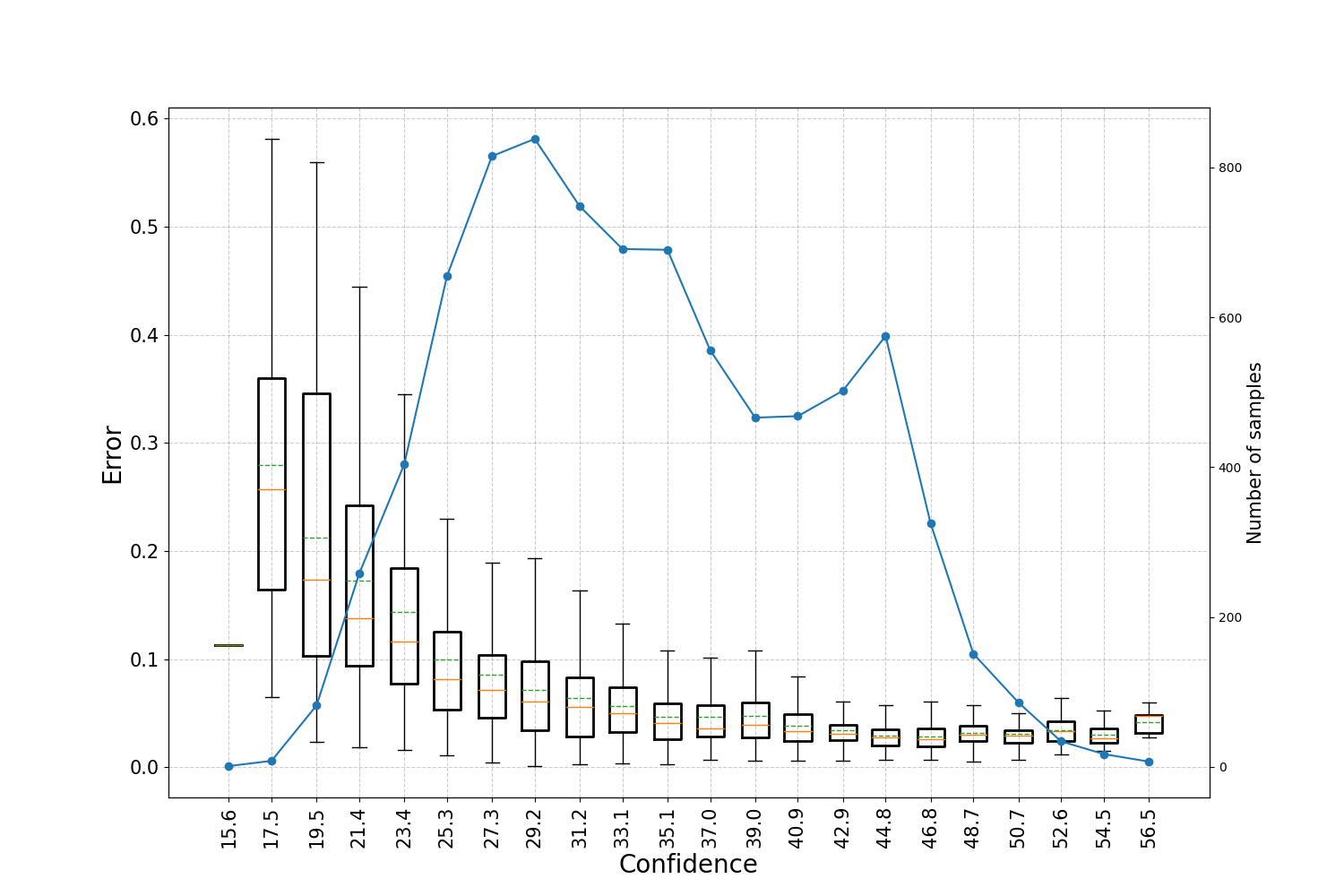} & 
        \includegraphics[trim={50px 15px 50px 50px}, clip, width=0.32\textwidth]{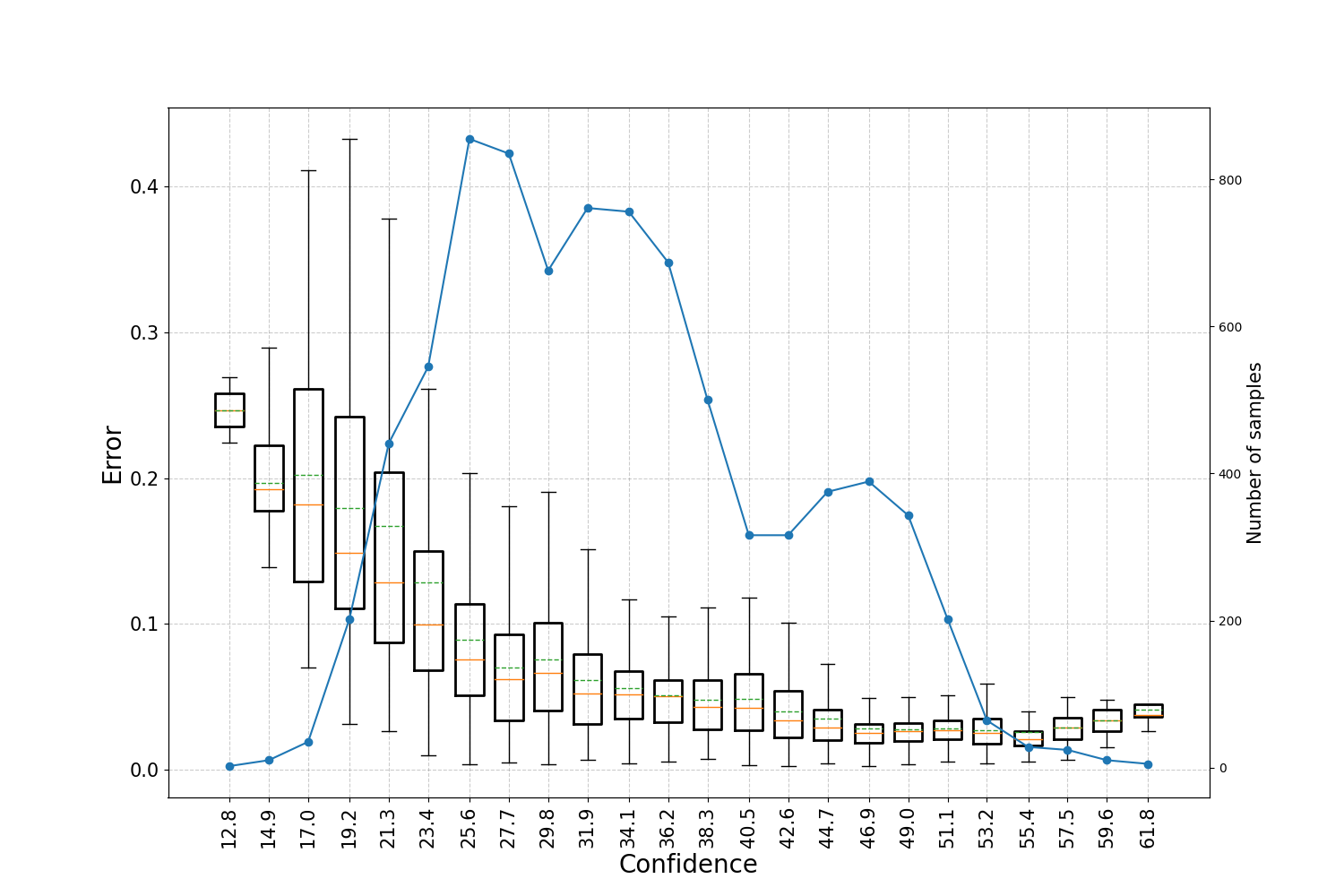} \\
        (a) All joints & (b) Left wrist & (c) Right wrist
    \end{tabular}
    \caption{Reliability diagrams for (a) all joints in the body, (b) left wrist joint, and (c) right wrist joint. Our model produces quite reliable confidence estimates, \ie, high confidence for predictions with low error and vice versa. The blue line, represented by the secondary y-axis on the right of each plot, shows the number of samples falling inside each bin. The mean of each box is illustrated with a dashed green line and the median with a solid orange line.}
    \label{fig:reliability}
\end{figure*}

To confirm this, we compare the average predicted concentration parameters for different axes of different joints, as illustrated in \cref{fig:dof_concentration}. 
These results show that the concentration parameters not only reflect the confidence of the network based on the input image, but also based on a learned representation of the human pose.

\paragraph{Relation between error and uncertainty.}

Uncertainty estimates around each joint should also convey how reliable the predictions are for that particular joint. 
This is an important aspect, as such uncertainty estimates may be used in multiple downstream tasks. 
Thus, it is essential that predicted uncertainties behave in line with expected error.

To evaluate the quality of estimated uncertainties (concentration parameters), we extend the reliability diagrams of \citet{guo2017calibration} and plot the relationship between estimated concentration parameters and MPJPE.
This is achieved by visualizing the trend of MPJPE as a function of concentration parameters.
To obtain this diagram for the test set, we divide the sum of the predicted $\mathcal{K}_i$s into $B$ equally-spaced interval bins, use the data points (pairs of MPJPE and $\mathcal{K}_i$ for each joint) that correspond to each interval, and compute a box plot for each.
This results in diagrams where the relation between the estimated confidences and produced error is visualized, see \cref{fig:reliability}.

We argue that a \emph{well-trained} model should produce low MPJPE when it makes a prediction with high confidence. 
Similarly, when the model predicts a joint with higher uncertainty, we expect to observe higher MPJPE. 
This is clearly illustrated in \cref{fig:reliability} when considering all joints together, as well as for two isolated joints\footnote{See the supplementary material for further analyses and reliability diagrams for all joints.}.

\subsection{Performance}
Our monocular and stereo networks run at 100 and 56 FPS respectively. 
This is twice as fast as their corresponding counterparts, xR-EgoPose and UnrealEgo, which run at 42 and 33 FPS.
Similarly, our model has only a 21.5M (monocular) and 21.9M (stereo) parameters, compared to 171.4M and 191.6M. 
This is mainly because we don't rely on predicting 2D heat-maps for each joint or using a two-stage network to get the final pose and joint location. 
All measurements were taken using an NVIDIA GTX-1080.


\section{Conclusions}
We have presented a solution for the problem of egocentric pose estimation from a head-mounted camera, motivated by the fact that current solutions: (1) predict only 3D joint locations, which reduces their utility for HMD applications, (2) present a complex architecture leading to a slower execution, (3) cannot capture the uncertainty of joints that are outside the image frame.

Our approach predicts matrix Fisher distributions over rotation matrices, from which we obtain the joint rotations, per-axis uncertainties, and recover the 3D joint positions. 
We demonstrate that uncertainty is correlated with error and so estimated uncertainties can be relied upon to explain the input image and predicted pose.
Qualitatively we notice, for example, that the uncertainty of a joint will be high if it is occluded or out of frame, and low otherwise.
We also show a high correlation between concentration parameters and degrees-of-freedom of a joint.
This indicates that our model can learn a representation of the range of motion of human joints, and so a prior on human pose.

As current egocentric datasets only include images taken from a fish-eye lens camera, which generally captures the entire body, we introduce the \SynthEgo{} dataset, a diverse photo-realistic synthetic dataset using a pinhole camera model.
This allows us to evaluate the quality of predicted poses from images with self-occlusions and non-visible body parts. 
Our method achieves better performance than the current state-of-the-art on this challenging dataset as well as generalizing well to real data.
Our approach is also twice as fast as previous methods and has $8\times$ fewer parameters.

\clearpage
{
    \small
    \bibliographystyle{ieeenat_fullname}
    \bibliography{main}
}

\end{document}


\maketitle

\section{Optimizing auxiliary tasks}

\begin{table}
\centering

\resizebox{\linewidth}{!}{%
\begin{tabular}{lccccccccc} 
\toprule
\multirow{2}{*}{Input} & Pose & Pose & SMPL-H* & 2D & Reprojection & Body & MPJPE & PA-MPJPE \\
& (rot. mat) & (Fisher) & Joint Loss & Landmarks & Error & Shape & (mm) & (mm) \\ 
\midrule
\multirow{7}{*}{Mono} & \checkmark            &               &                 &              &                                                     &            & 160.66                         & 133.36                             \\
                && \checkmark    &                 &              &                                                     &            & 146.53                         & 121.86                             \\
                && \checkmark  & \checkmark    &              &                                                     &            & 118.39                         & 98.00                              \\
                && \checkmark  & \checkmark    & \checkmark &                                                    &            & 116.74                         & 96.24                              \\
                && \checkmark  & \checkmark    & \checkmark & \checkmark                                        &            & 116.43                         & 96.18                              \\
                && \checkmark  & \checkmark    & \checkmark & \checkmark                                        & \checkmark       & 121.97                         & 96.96                              \\
                && \checkmark  &   \checkmark               &              &                                                     & \checkmark       & 123.56                         & 98.93                              \\
\hline
\multirow{7}{*}{Stereo} & \checkmark      &               &                 &              &                    &            & 148.37                         & 123.95                             \\
                && \checkmark    &                 &              &                    &            & 139.37                         & 115.67                             \\
                && \checkmark    & \checkmark      &              &                    &            & 110.01                         & 90.68                              \\
                && \checkmark    & \checkmark      & \checkmark   &                    &            & 110.52                         & 90.67                              \\
                && \checkmark    & \checkmark      & \checkmark   & \checkmark         &            & 111.74                         & 92.09                              \\
                && \checkmark    & \checkmark      & \checkmark   & \checkmark         & \checkmark & 116.06                         & 92.17                             \\
                && \checkmark    & \checkmark      &              &                    & \checkmark & 115.10                         & 91.79                              \\
\bottomrule
\end{tabular}%
}

\caption{Ablation study of different auxiliary predictions and losses using monocular and stereo inputs on the synthetic test set. For columns 2-7. (2) The network predicts rotation matrices, (3) The network predicts Fisher parameters. (4) Usage of SMPL-H* differentiable function to optimize 3D joint locations. (5) 2D landmarks predicted as auxiliary output (6) Reprojection error as loss. (7) Body identity, $\boldsymbol\beta$, prediction as extra output.}
\label{tab:ablation}

\end{table}

We trained the network using a variety of auxiliary tasks like predicting 2D landmarks, body shape and minimizing the re-projection error, results are shown in \cref{tab:ablation} and described below.

Overall the performance of the network is essentially unchanged when predicting 2D landmarks on the image, and minimizing the re-projection error worsens the predictions.
Possibly these tasks are made more challenging due to the extreme viewpoint of the camera.

\paragraph{2D landmarks loss.}
To predict 2D landmarks as an auxiliary task, we used the approach of \citet{wood2022dense}, predicting probabilistic 2D joints as a 2D Gaussian with mean $\mathbf{J}_{2D_i}=\{\mu_{x_i}, \mu_{y_i}\}$ and covariance $\sigma_i \mathbf{I}$. 
We train it using Gaussian negative log likelihood (GNLL) and the visible ground truth targets $\mathcal{L}_{2D} =  \sum^{N}_{i=i} v_i\left (\log(\sigma_i) + \frac{\left \|(\mathbf{\hat{J}}_{2D_i}- \mathbf{J}_{2D_i} ) \right \|^2}{2\sigma^2_i} \right )$, where $v_i$ is $1$ if the joint $i$ is inside the image frame and $0$ otherwise. 
We only predict the 2D landmarks on the left camera. 

\paragraph{Body identity loss.}
We add the ability to predict body shape parameters, $\boldsymbol\beta$, to the network in some experiments. 
This prediction is trained with negative log-likelihood loss over the distribution of the body shape parameters $\mathcal{L_{\boldsymbol\beta}}=-log(\mathcal{N}(\boldsymbol\beta; \mu_\beta(\textbf{X}), \text{diag}(\sigma(\textbf{X})^2_\beta)))$~\cite{sengupta2021hierarchical, kendall2018multi}.
We then are able to use this output instead of ground-truth shape parameters when regressing predicted joint locations.

\paragraph{Reprojection error.}
For external view methods~\cite{choutas2020monocular, sengupta2021hierarchical, pavlakos2018learning, omran2018neural, varol2018bodynet, kolotouros2019learning}, it is common to assume an orthographic projection. 
However, due to the location of the egocentric cameras, this assumption cannot be followed, as the upper-body is closer to the camera than the feet. 
Therefore, a perspective projection is assumed instead.

We convert the predicted 3D joint positions into 2D locations by projecting them onto the image plane using the camera parameters, $\mathbf{\hat{J}}'_{2D}=\mathbf{C}[\mathbf{R}|\mathbf{t}]\mathbf{\hat{J}}_{3D}$, where $\mathbf{C}$ are the intrinsic camera parameters and $[\mathbf{R}|\mathbf{t}]$ are the rotation and translation of the camera. 
Then, we minimize the L2 distance between the 2D joints and visible ground truth joints $\mathcal{L}_{reproj}=\sum_{i=1}^{N}\left \| v_i(\mathbf{\hat{J}}'_{2D_i}- \mathbf{J}_{2D_i} ) \right \|^2_2$. 
We only reproject the 3D joints to the left image, future work could use the right image as well to improve the predictions~\cite{akada2022unrealego}.

\section{Per-vertex position uncertainty}

With the matrix Fisher distribution over the joint rotations, we can obtain a per-vertex uncertainty which reflects how far a body part is from the mean pose in Euclidean space. 
For this, we use the rejection sampler of \citet{sengupta2021hierarchical} to sample 50 SMPL-H* meshes from the predicted distribution.
We then obtain the average Euclidean distance from the sample mean for each vertex over all the samples and compute the standard deviation of each vertex. 
This type of uncertainty is important, as it takes into consideration the kinematic tree of the body, given that the rotation of the parent joints will affect the final position of their children and vertices that correspond to them. 

\section{Real dataset collection}

\begin{figure}
    \centering
    \includegraphics[width=\linewidth]{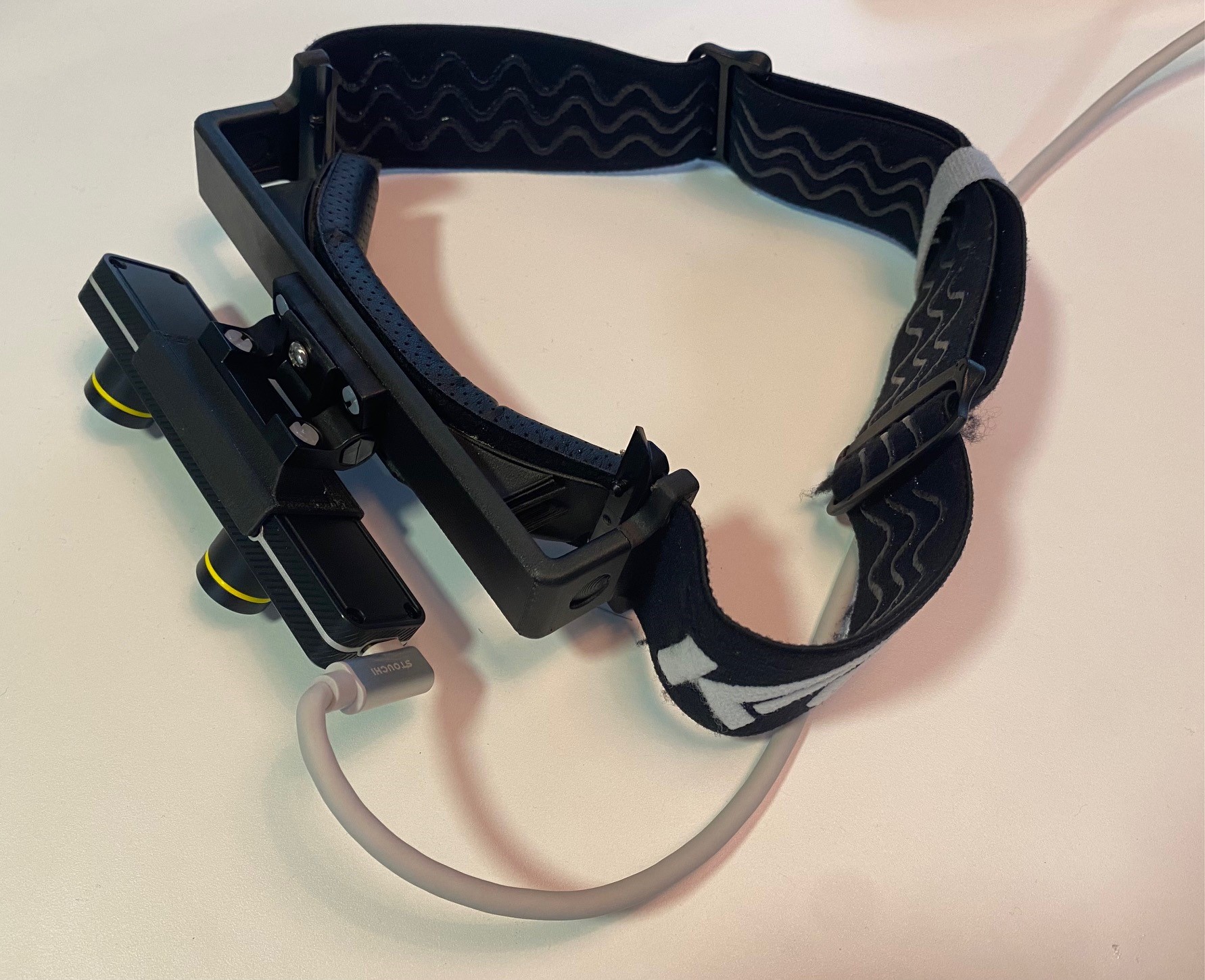}
    \caption{Headset and stereo camera used for egocentric image capture during data collection.}
    \label{fig:headset}
\end{figure}
\begin{figure}
    \centering
    \includegraphics[width=\linewidth]{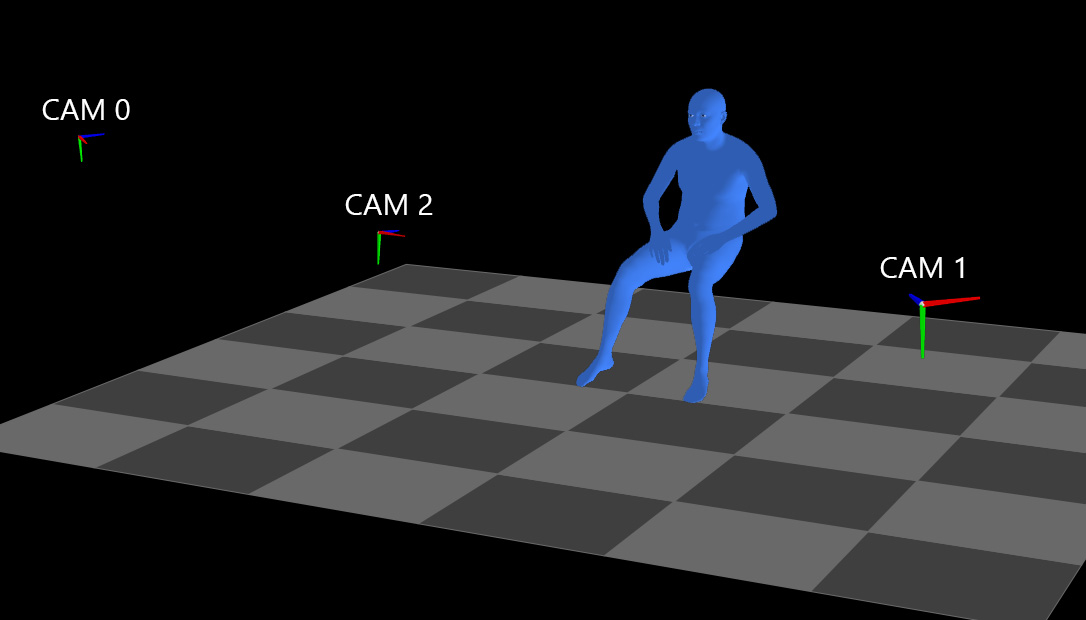}
    \caption{Visualization of external-view capture rig showing three cameras arranged around the subject.}
    \label{fig:recording_rig}
\end{figure}

To collect our real data we have participants wear a ZED mini stereo camera\footnote{\url{https://www.stereolabs.com/}} on a head-mounted rig as shown in \cref{fig:headset}.
While wearing this, the participant stands in a capture volume with three hardware synchronized Azure Kinect cameras positioned around them as shown in \cref{fig:recording_rig}.
The participant is then directed to carry out certain gestures or motions while both devices are recording.
The participant starts the sequence by clapping to provide a synchronization signal between the two (egocentric and external) video sources.

\begin{figure}
    \centering
    \includegraphics[height=0.268\linewidth]{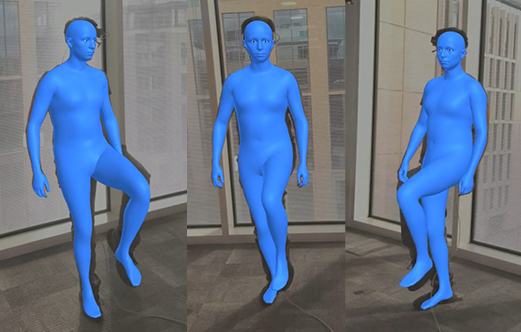}
    \hfill
    \includegraphics[height=0.268\linewidth]{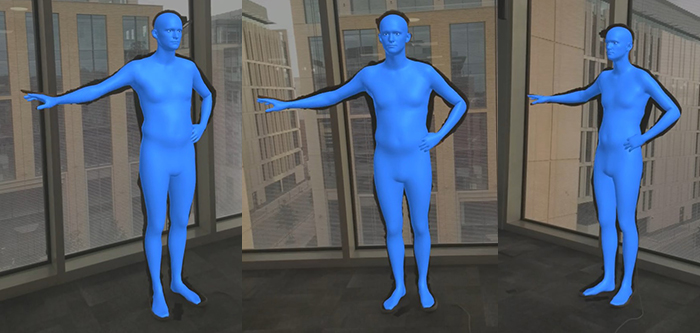}
    \\
    \includegraphics[height=0.268\linewidth]{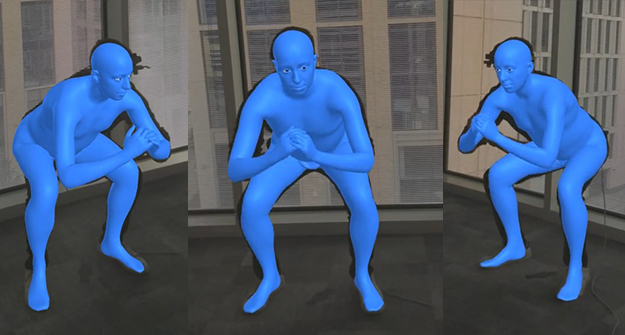}
    \hfill
    \includegraphics[height=0.268\linewidth]{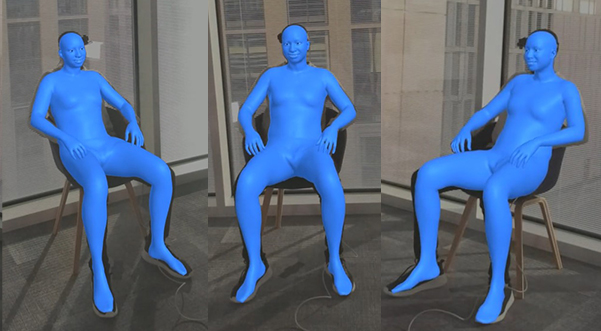}
    \caption{Example frames from our real dataset collection showing three external views used to obtain ground-truth pose data with fitting result overlaid.}
    \label{fig:real_data}
\end{figure}

The egocentric data is used as input to the models described in the main paper.
The external views are used to obtain ground-truth body shape and pose data using the procedure outlined by \citet{tech_report}.
Some ground-truth registration results are shown in \cref{fig:real_data}.

We record 11 participants carrying out 62 sequences in total.
We manually filter these based on registration quality to obtain a final dataset of 8378 frames.

\section{Additional results}
\begin{figure*}[p]
\begin{center}
   \includegraphics[width=\linewidth]{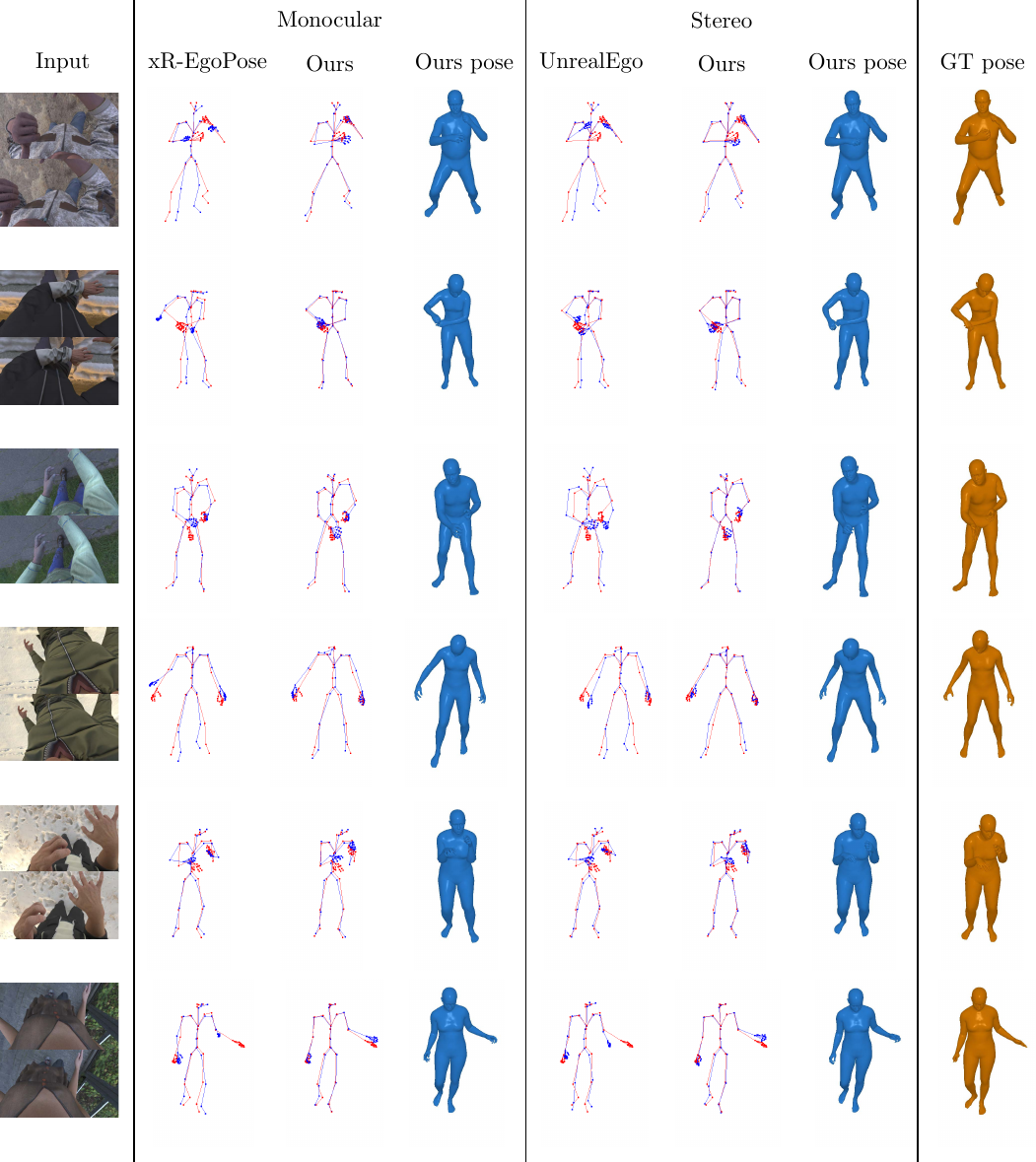}
\end{center}
   \caption{Extension of 3D joint location comparison on synthetic data. The input column has the left (top) and right (bottom) images, the monocular networks use only the left. The predicted joint positions (blue) are overlaid on the ground-truth (red). We also show the predicted (blue) and GT (orange) body meshes.}
\label{fig:output_preds_sx_extend}
\end{figure*}

\begin{figure*}[p]
\begin{center}
   \includegraphics[width=\linewidth]{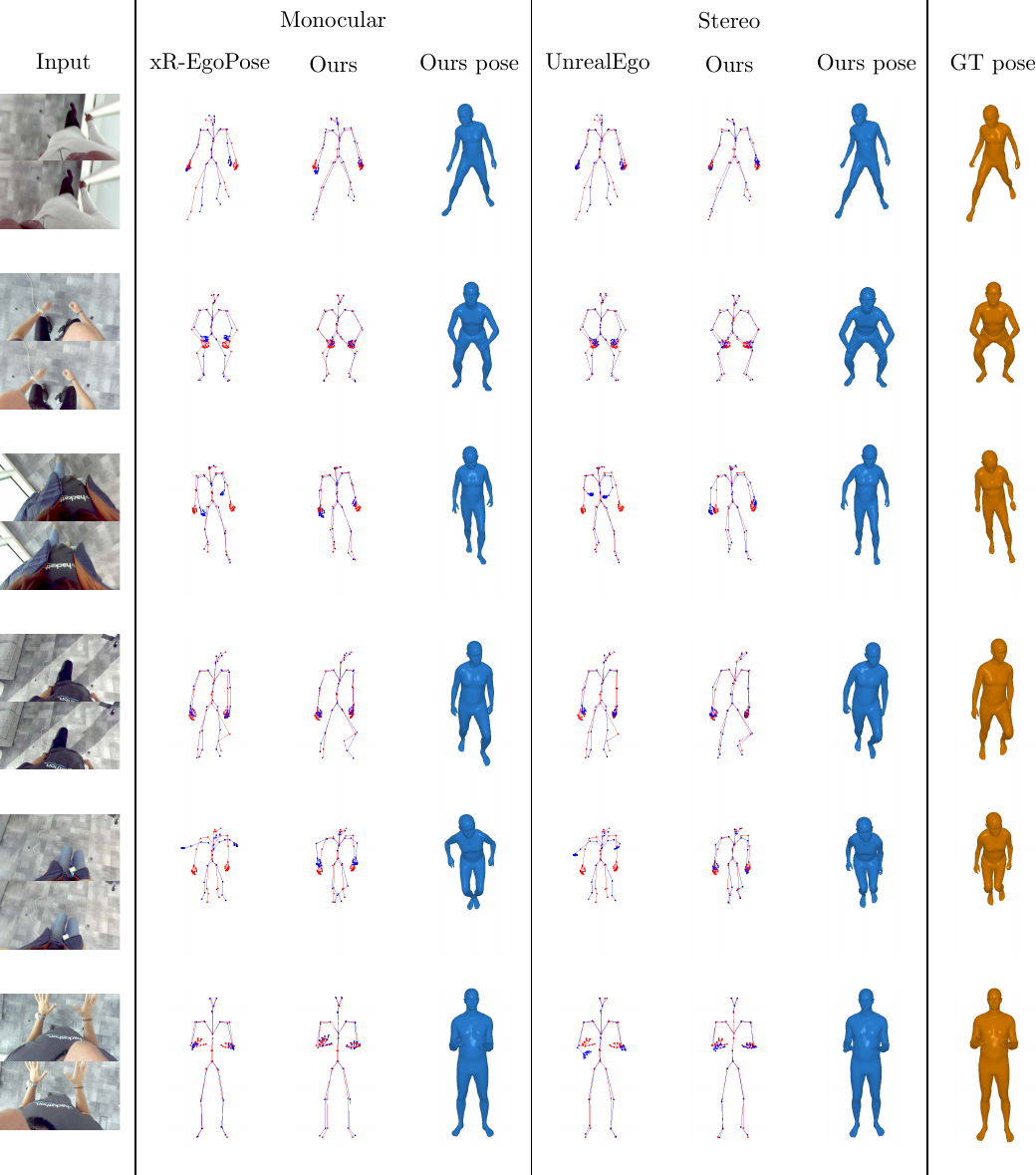}
\end{center}
   \caption{Extension of 3D joint location comparison on real data. }
\label{fig:output_preds_extend}
\end{figure*}

\paragraph{Comparison to state-of-the-art.}
We extend the comparisons to recent work~\cite{akada2022unrealego,tome2019xr} from the main paper in \cref{fig:output_preds_sx_extend,fig:output_preds_extend}. 
The former shows more results on our synthetic data, and the latter on real data.

\begin{figure*}
    \centering
    \includegraphics[width=\linewidth]{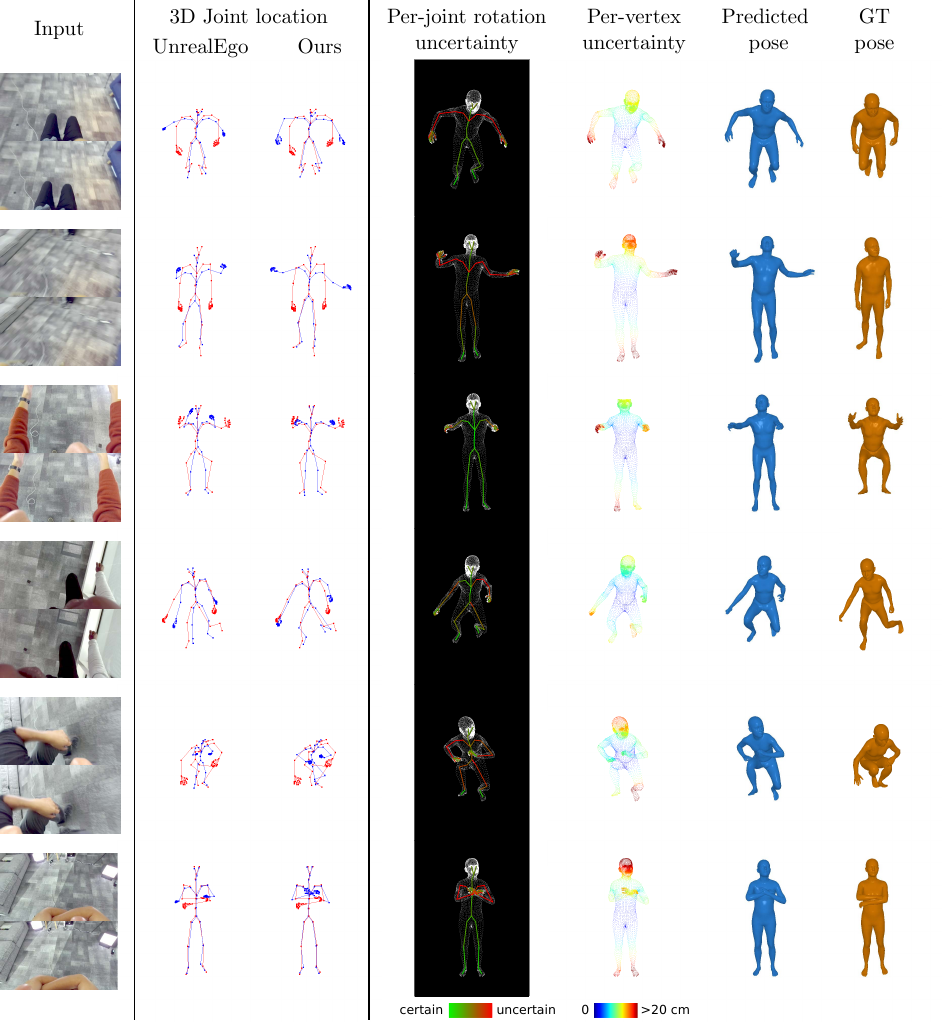}
    \caption{Challenging cases and uncertainty prediction. We show the predicted 3D joint locations of UnrealEgo and our method on some challenging inputs. It is observed that even when some joints are predicted wrongly, our network is not confident about them and outputs a high uncertainty score (fourth column).}
\label{fig:uncertainty_pred_extend}
\end{figure*}

\paragraph{Challenging cases and uncertainty prediction.}
We analyze the failure cases of our network. 
We observe that a whole kinematic chain typically has to be missing in the image for the network to predict a completely incorrect expected pose. 
For example, if the joints in the arm and shoulder are missing the network will predict incorrect expected pose for these parts, see first and second row in \cref{fig:uncertainty_pred_extend}. 
However, in these cases the network assigns low certainty values to those joints, a highly desirable behavior.
This is understandable as the network loses all possible information about those extremities, which makes their location ambiguous.
However, if part of this chain is seen, like the arms in the third row, the network is able to use this limited information to infer broadly correct poses.

We also observe that if the legs are not at all visible (from hip to feet), the network will rely on a learned prior and predict a standing position (see third row of \cref{fig:uncertainty_pred_extend}).
It seems in this case that the prior is strong enough that the network predicts the lower body pose with high confidence, despite there being no observable information to support this.
This is not a desirable behavior and might be addressed by incorporating a more diverse set of poses in the training data.

\paragraph{Joint degrees-of-freedom.}

\begin{figure*}[p]
    \centering
    \includegraphics[width=.8\textwidth]{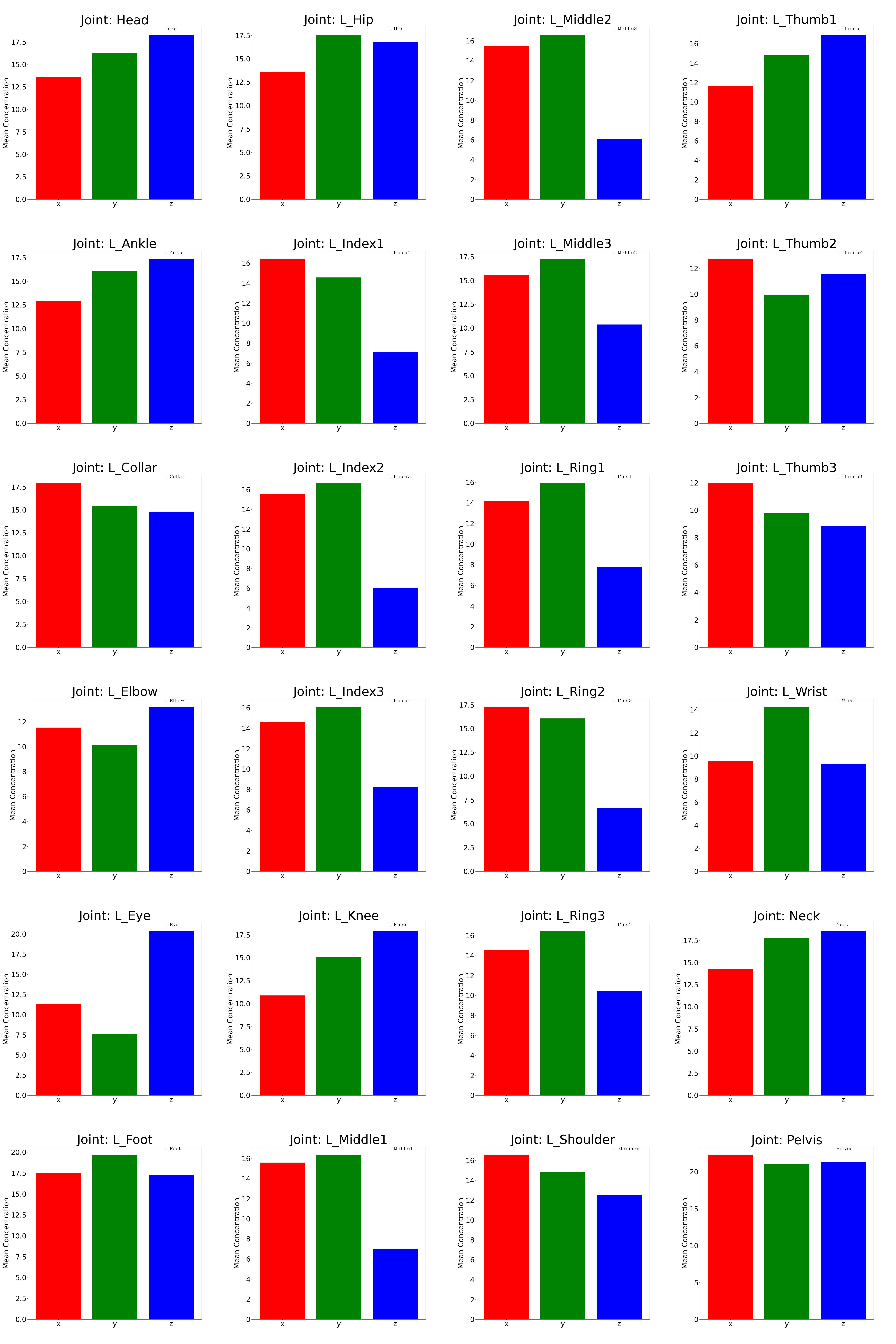}
    \caption{Concentration parameter and joint degrees-of-freedom
relationships. Averaged concentration parameters from the real
dataset show that higher concentration values are correlated with
axes with lower freedom.}
    \label{fig:axes_fov}
\end{figure*}

\cref{fig:axes_fov} shows plots of concentration parameters per-axis for 30 of the 54 joints in our body model, we exclude the right side for symmetrical joints.
These values are obtained by averaging over predictions on our real dataset.
We observe that higher concentration values correlate with axes with lower freedom.

\paragraph{Reliability diagrams.}

\begin{figure*}[p]
    \centering
    \includegraphics[width=\textwidth]{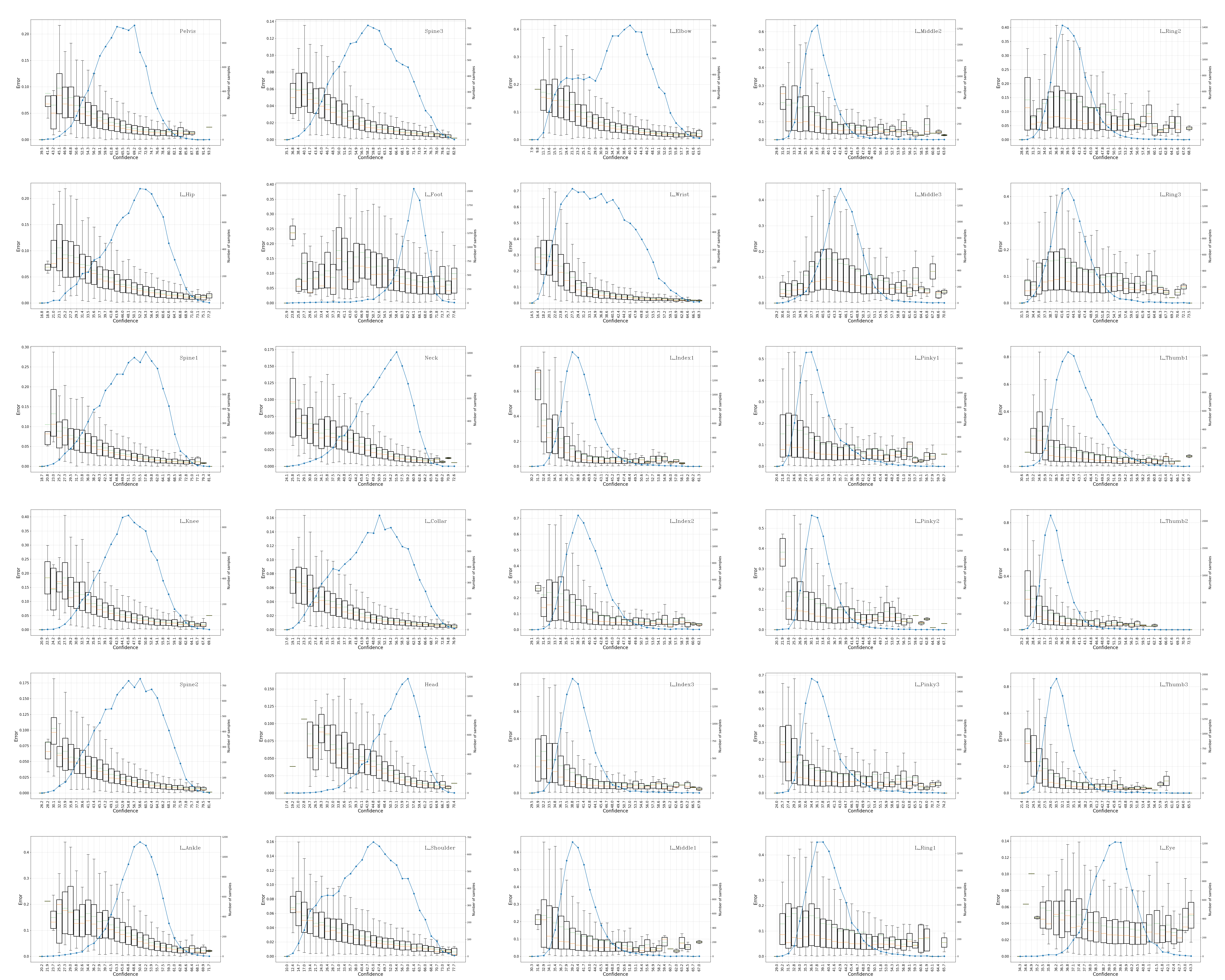}
    \caption{Reliability diagrams for 30  joints. Plots are generated from synthetic data and monocular view.}
    \label{fig:reliability_synthetic_mono}
\end{figure*}

\begin{figure*}[p]
    \centering
    \includegraphics[width=\textwidth]{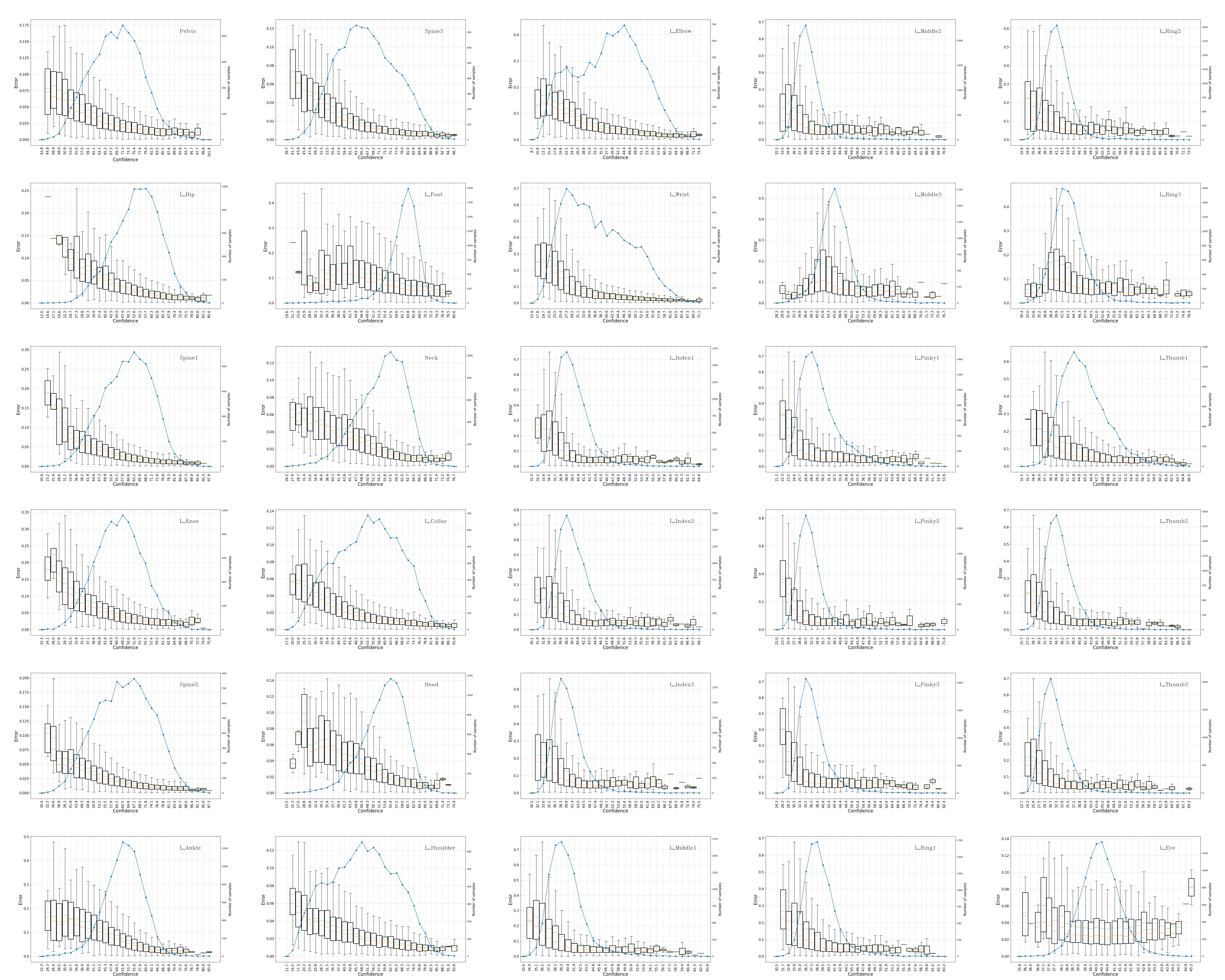}
    \caption{Reliability diagrams for 30  joints. Plots are generated from synthetic data and stereo view.}
    \label{fig:reliability_synthetic_stereo}
\end{figure*}

\begin{figure*}[p]
    \centering
    \includegraphics[width=\textwidth]{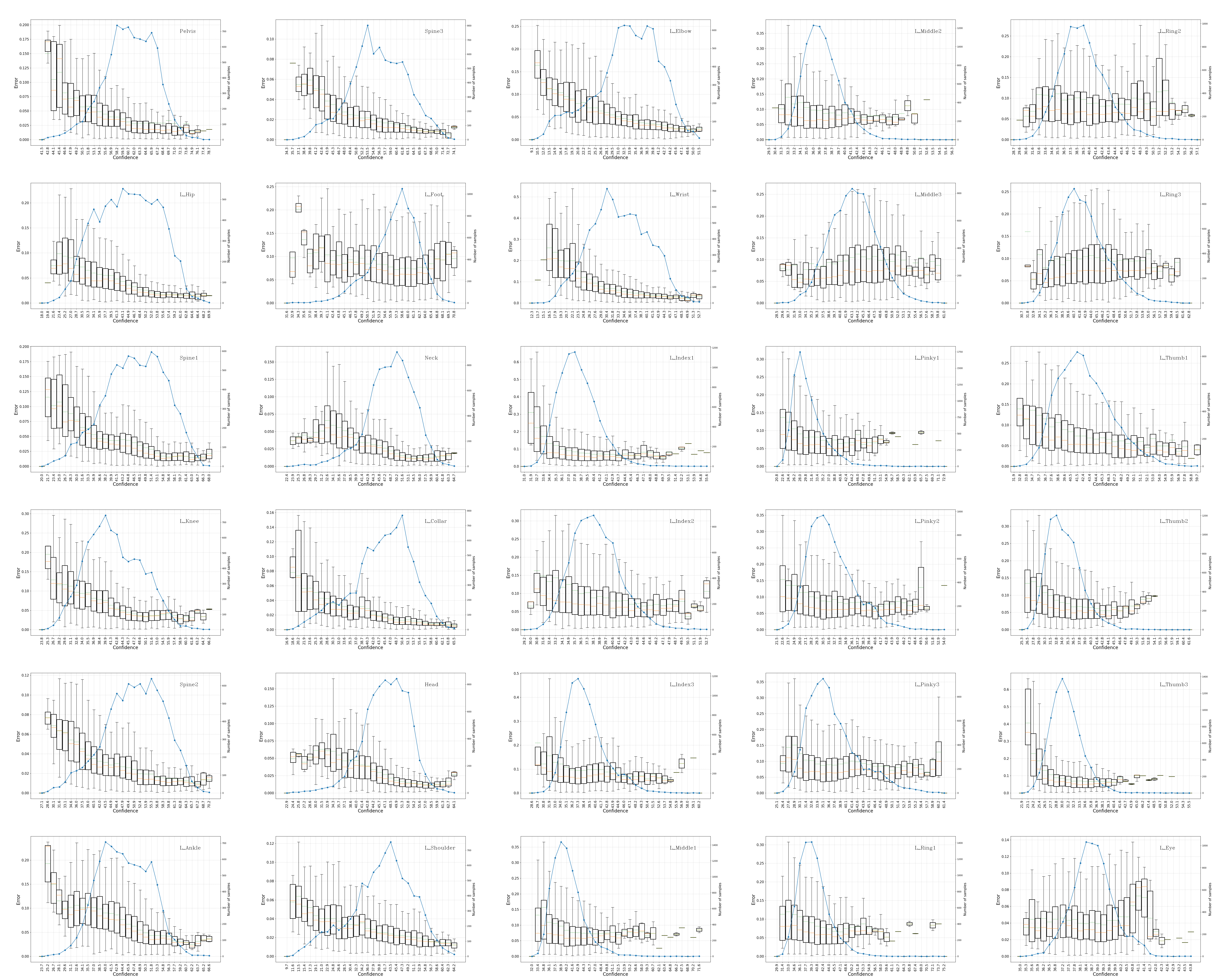}
    \caption{Reliability diagrams for 30  joints. Plots are generated from real data and monocular view.}
    \label{fig:reliability_real_mono}
\end{figure*}

\begin{figure*}[p]
    \centering
    \includegraphics[width=\textwidth]{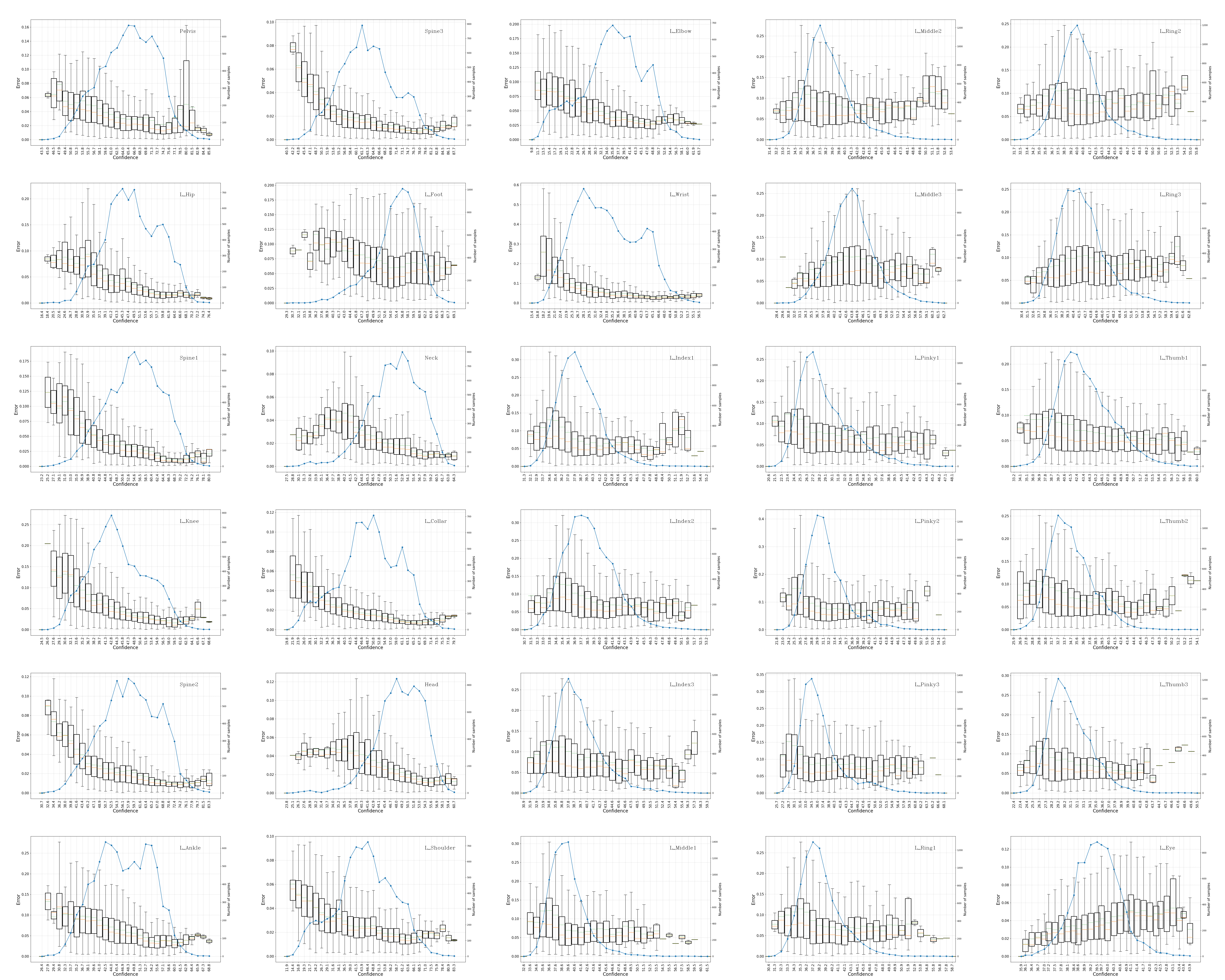}
    \caption{Reliability diagrams for 30 joints. Plots are generated from real data and stereo view.}
    \label{fig:reliability_real_stereo}
\end{figure*}

\cref{fig:reliability_synthetic_mono,fig:reliability_synthetic_stereo,fig:reliability_real_mono,fig:reliability_real_stereo} show reliability diagrams for 30 of the 54 joints in our body model, we exclude the right side for symmetrical joints. 
We plot these diagrams for synthetic and real data in the monocular and stereo cases.
These plots demonstrate the strong correlation between error and uncertainty for almost all joints in the skeleton.
For finger joints the relation is less pronounced, this is likely because hand pose is particularly challenging to predict, particularly from a single full-body region-of-interest (ROI).
Future work could use separate ROIs for each hand to achieve very high quality predictions~\cite{choutas2020monocular}.

\section{\SynthEgo{} dataset statistics}

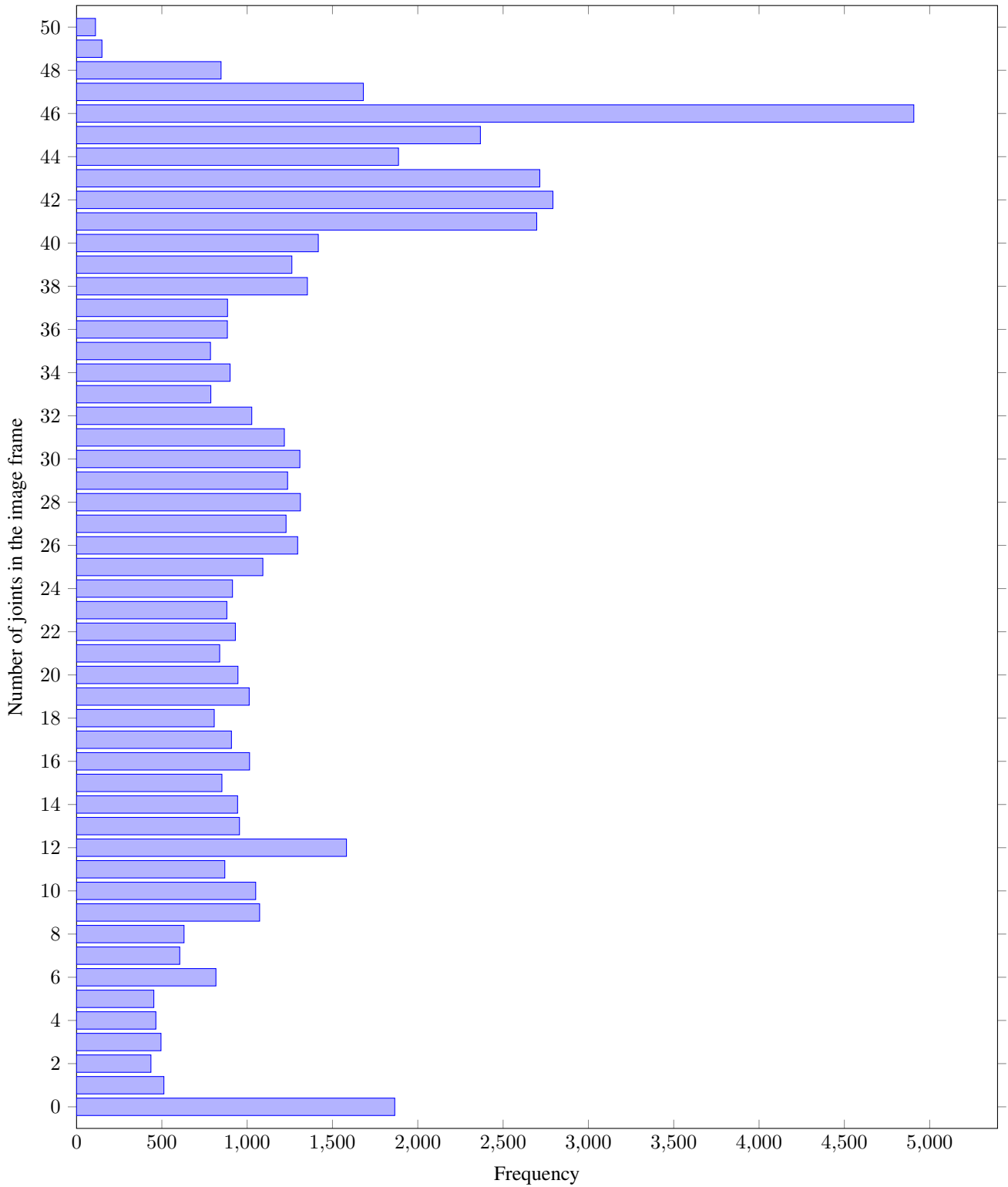
\begin{figure*}[p]
    \centering
\begin{tikzpicture}
 
\begin{axis}[xbar, width=\textwidth, height=1.2\textwidth, bar width=0.3cm, ymin=-1, ymax=51, xmin=0, ylabel=Number of joints in the image frame, xlabel=Frequency]
\addplot coordinates {
(1865,0)
(512,1)
(436,2)
(495,3)
(465,4)
(453,5)
(817,6)
(605,7)
(630,8)
(1073,9)
(1050,10)
(869,11)
(1582,12)
(955,13)
(944,14)
(852,15)
(1014,16)
(908,17)
(807,18)
(1012,19)
(946,20)
(839,21)
(931,22)
(881,23)
(914,24)
(1092,25)
(1296,26)
(1228,27)
(1312,28)
(1237,29)
(1309,30)
(1218,31)
(1027,32)
(787,33)
(900,34)
(785,35)
(884,36)
(885,37)
(1353,38)
(1262,39)
(1417,40)
(2697,41)
(2792,42)
(2715,43)
(1887,44)
(2367,45)
(4907,46)
(1681,47)
(847,48)
(149,49)
(111,50)
};
\end{axis}
 
\end{tikzpicture}
    
    \caption{Histogram of the number of joints within the image frame in our synthetic dataset. There is a wide range in the number of visible joints, including some samples with no visible joints, \eg, where the user is looking upwards. Note that self-occlusion is not accounted for in these measurements.}
    \label{fig:joint_vis}
\end{figure*}

\cref{fig:joint_vis} shows a histogram of the total number of joints within the image frame for the \SynthEgo{} dataset.
This analysis does not include the effect of self-occlusion on visibility.
There is significant variety in terms of visibility of the body, including some frames where no joints are visible.
These are likely situations where the person is looking directly upwards, for example. 
We also compute the MPJPE and PA-MPJPE for the cases where the joints are out-of-frame and in-frame. 
For both cases our method performs better than previous state-of-the-art methods, see \cref{tab:joint_in-out-vis_comp}.

\begin{table}
\centering
\footnotesize
\resizebox{\linewidth}{!}{
\begin{tabular}{llcccc} 
\toprule
\multicolumn{1}{c}{\multirow{2}{*}{Input}} & \multicolumn{1}{c}{\multirow{2}{*}{Method}} & \multicolumn{2}{c}{Out-of-frame joints } & \multicolumn{2}{c}{In-frame joints }  \\ 
\cline{3-4} \cline{5-6}
\multicolumn{1}{c}{}                       & \multicolumn{1}{c}{}                        & MPJPE           & PA-MPJPE               & MPJPE          & PA-MPJPE             \\ 
\midrule
\multirow{2}{*}{Monocular}                 & xR-EgoPose                                  & 203.77          & 173.06                 & 115.35         & 102.96               \\
                                           & Ours                                        & \textbf{166.36} & \textbf{136.81}        & \textbf{76.58} & \textbf{64.17}       \\ 
\midrule
\multirow{2}{*}{Stereo}                    & UnrealEgo                                   & 194.03          & 166.81                 & 98.07          & 91.90                \\
                                           & Ours                                        & \textbf{155.64} & \textbf{127.46}        & \textbf{70.24} & \textbf{58.63}       \\
\bottomrule
\end{tabular}
}
\caption{MPJPE (mm) comparison with previous methods for out-of-frame and in-frame joints on the \SynthEgo{} dataset. Our method significantly outperforms previous methods in both cases.}
\label{tab:joint_in-out-vis_comp}
\end{table}


{
    \small
    \bibliographystyle{ieeenat_fullname}
    \bibliography{main}
}